\documentclass[journal]{IEEEtran}
\usepackage[utf8]{inputenc}
\usepackage[table]{xcolor}
\usepackage{tikz}
\usepackage[noadjust]{cite}

\usetikzlibrary{matrix,shapes.geometric, arrows.meta,positioning,fit, spy, intersections, arrows,shapes.multipart, calc}
\tikzstyle{line}=[draw]
\usepackage{kpfonts}
\usepackage{accents}
\usepackage{pgfplots}
\usepackage[ruled,vlined]{algorithm2e}
\usepackage{algcompatible}
\usepgfplotslibrary{patchplots}
\pgfplotsset{compat=newest}
\usepgfplotslibrary{fillbetween}
\usepackage{MnSymbol,wasysym}
\usepackage{subfig}
\usepackage{multirow,bigstrut}
\usepackage{enumitem}
\usepackage[nomessages]{fp}
\usepackage{footnote}
\usepackage{adjustbox}
\usepackage{arydshln}
\usepackage{dblfloatfix} 

\usepackage{color,soul}
\usepackage{amsthm}
\usepackage{amsmath}
\newtheorem{definition}{Definition}[section]
\newtheorem{theorem}{Theorem}[section]
\newtheorem{proposition}{Proposition}[theorem]
\renewcommand{\textcolor}[2]{#2}

\definecolor{mahogany}{rgb}{0.75, 0.25, 0.0}
\definecolor{pgreen}{rgb}{0.0, 0.4, 0.0}
\definecolor{persianblue}{rgb}{0.11, 0.22, 0.73}
\definecolor{darkcyan}{rgb}{0.0, 0.55, 0.55}
\definecolor{cadmiumred}{rgb}{0.89, 0.0, 0.13}
\definecolor{amber}{rgb}{1.0, 0.75, 0.0}
\pgfmathdeclarefunction{gauss}{3}{\pgfmathparse{(1*#3)/(#2*sqrt(2*pi))*exp(-((x-#1)^2)/(2*#2^2))}}%
\usepackage{xparse,xfp}
\NewExpandableDocumentCommand{\minimum}{m}{%
  \fpeval{min(#1)}%
}
\title{\textbf{A Temporal Type-2 Fuzzy System for Time-dependent Explainable Artificial Intelligence}}
\author{Mehrin~Kiani, %~\IEEEmembership{Student Member,~IEEE,}
        Javier~Andreu-Perez$^*$,~\IEEEmembership{Senior Member,~IEEE,}
        and~Hani~Hagras,~\IEEEmembership{Fellow,~IEEE}% <-this % stops a space
\thanks{$^*$ Corresponding author: {\fontfamily{cmtt}\selectfont javier.andreu@essex.ac.uk}}
\thanks{M. Kiani, J. Andreu-Perez and H. Hagras are with the School
of Computer and Electronic Engineering, University of Essex,
Colchester, CO4 3SQ, United Kingdom.}}% <-this % stops a space
\date{}
\makeatletter
\newcommand\Xsubsubsection{\@startsection{subsubsection}{3}{\z@}%
                                     {-3.25ex\@plus -1ex \@minus -.2ex}%
                                     {1.5ex \@plus .2ex}%
                                     {\normalsize\leftskip 0ex}}
\renewcommand\subsubsection[1]{\Xsubsubsection{#1}\leftskip 0ex}
\makeatother
\newcommand{\midarrow}{\tikz \draw[->, ultra thick, rotate = -90] (0,0) -- +(.1,0);}
\newcommand{\midarrowother}{\tikz \draw[->, ultra thick, rotate = 90] (0,0) -- +(.1,0);}
\usepackage{etoolbox}
\AtBeginEnvironment{algorithm}{\algoequations}
\AtEndEnvironment{algorithm}{\restoreequations}
\newcounter{algosavedequation}
\newcounter{newAlgo}
\newcommand{\algoequations}{%
  \setcounter{algosavedequation}{\value{equation}}%
  \setcounter{equation}{0}%
  \renewcommand{\theequation}{Alg \arabic{newAlgo}.\arabic{equation}}%
}
\newcommand{\restoreequations}{%
  \setcounter{equation}{\value{algosavedequation}}%
}

\begin{document}

\maketitle
% ---------------------------------------------------------------------------------
%---------------------------------------------------------------------
\begin{abstract}
%---------------------------------------------------------------------
%---------------------------------------------------------------------
%Fuzzy systems are used for modelling real-life processes characterised with uncertainties. 
%The capability of explainable artificial intelligence (XAI) systems to compute with words and integrate uncertainty in given crisp measurements renders them particularly suitable for explainable modelling of real-life processes.
%--------------------------------------
%Background
%--------------------------------------
%Prof sentence suggestion
Explainable Artificial Intelligence (XAI) is a paradigm that delivers transparent models and decisions, which are easy to understand, analyze, and augment by a non-technical audience. Fuzzy Logic Systems (FLS) \textcolor{blue}{based XAI} can provide \textcolor{blue}{an explainable} framework, while also modeling  uncertainties present in real-world environments, which renders \textcolor{blue}{it} suitable for applications where explainability is a requirement. However, most real-life processes are not characterized by high levels of uncertainties alone; they are inherently time-dependent as well, i.e., the processes change with time.
%--------------------------------------
%Aim
%--------------------------------------
To account for the temporal component associated with a process, in this work, we present novel \emph{Temporal Type-2 FLS Based Approach} for time-dependent XAI (TXAI) systems, which can account for the likelihood of a measurement's occurrence in the time domain using (the measurement's) frequency of occurrence. 
%--------------------------------------
%Method
%--------------------------------------
In Temporal Type-2 Fuzzy Sets (TT2FSs), a four-dimensional (4D) time-dependent membership function is developed where relations are used to construct the inter-relations between the elements of the universe of discourse and its frequency of occurrence. 
%The two main advantages of using TXAI systems over standard XAI systems are: 1) TXAI systems can account for the likelihood of a measurement's occurrence in the time universe using (the measurement's) frequency of occurrence % (i.e., utilizing the associated time information to seek confidence in the given measurement), 
%and 2) TXAI systems can delineate possible trajectories of a temporal process with respect to time using the frequency of occurrence values embedded in the TXAI model.
%--------------------------------------
%Results
%--------------------------------------
The proposed TXAI system with TT2FSs is exemplified with a step-by-step numerical example and an empirical study using a real-life intelligent environments dataset to solve a time-dependent classification problem (predict whether or not a room is occupied depending on the sensors readings at a particular time of day). The TXAI system performance is also compared with other state-of-the-art classification methods with varying levels of explainability. The TXAI system manifested better classification prowess, with 10-fold test datasets, with a mean recall of 95.40\% than a standard XAI system (based on non-temporal general type-2 (GT2) fuzzy sets) that had a mean recall of 87.04\%. TXAI also performed significantly better than most non-explainable AI systems between 3.95\%, to 19.04\% improvement gain in mean recall. Temporal convolution network (TCN) was marginally better than TXAI (by 1.98\% mean recall improvement) although with a major computational complexity. In addition, TXAI can also outline the most likely time-dependent trajectories using the frequency of occurrence values embedded in the TXAI model; viz. given a rule at a determined time interval, what will be the next most likely rule at a subsequent time interval.
%--------------------------------------
%Impact
%--------------------------------------
In this regard, the proposed TXAI system can have profound implications for delineating %and throw light on 
\textcolor{blue}{the evolution of }real-life time-dependent processes, such as behavioural or biological processes.
%---------------------------------------------------------------------
%---------------------------------------------------------------------
\end{abstract}
%---------------------------------------------------------------------
% ---------------------------------------------------------------------------------
\section{Introduction} \label{sec:Intro}
%---------------------------------------------------------------------
%---------------------------------------------------------------------
%brain computer interfaces \cite{AndreuPerez_2016_T2}
% %speech recognition \cite{Alhawiti_2015_SpeechRecAI}
% risk management \cite{Baryannis_2019_RiskMgtAI}
% decision making \cite{Wallace_2016} \cite{Hussein_2020_DecisionMaking}
%risk management \cite{Bussmann_2021}
%, smart cities \cite{Allam_2019_SmartCitiesAI} 
Over the last few decades, the widespread application of artificial intelligence (AI) systems have enhanced many aspects of everyday life from risk management \cite{Lee_2021}, sky shepherding of sheep \cite{Yaxley_Abbass_2021}, medical image segmentation \cite{Li_Wagner_2016}, recognition of expertise level \cite{Kiani_2019_EFCM}, 
mobile applications \cite{Lu_2018_BrainIntelligenceAI} to Covid-19 detection based on cough samples \cite{AndreuPerez_2021_CovidDetection}. Although opaque AI systems offer remarkable prediction accuracy, they are limited by a lack of explanation behind their predictions. A lack of explanation renders the AI systems untrustworthy, and particularly inapplicable where users want to understand the decision process of the AI system. To this end, there is a growing need for transparent, human-understandable AI systems called explainable AI (XAI) systems \cite{Hagras_2018}. Several approaches taken towards the development of XAI systems include: 1) Intrinsic: a method in which model inference structure is fully transparent \textcolor{blue}{such as short decision trees or sparse linear models}, and 2) Post-hoc: a model-agnostic meta-model is used to decipher the inference rationale of a black-box model \textcolor{blue}{permutation feature importance can be computed for decision trees}. Within post-hoc methods attempts to unravel a black-box model into a surrogate intrinsic model have also been undertaken. A particular category of these are the anchor-based models. 

\input{Figures_texFiles_Used/Types_Of_FuzzySets}

Although anchor-based approach provides a step towards implementing human-understandable explanations \cite{alicioglu2021}, explanatory patterns rest on hard thresholds and are constrained by Boolean logic. \textcolor{blue}{ However, real-life processes are characterised with uncertainty and therefore hard thresholds based models are not particular well-suited to model them (real-life processes)}. In this regard, another approach to implement XAI systems is fuzzy logic systems (FLS) \cite{Hagras_2018,Zadeh_1975}. The FLS based XAI systems are well-suited for explainable modelling of real-life processes because of FLS capability to handle uncertainty in the input data, and subsequently improve the process model and performance. In addition, the use of conceptual labels (CoLs) that model uncertainty and axioms of FLS based XAI systems pave way for human-understandable models for describing complex, real-life processes.

%The explainable artificial intelligence (XAI) systems \cite{Hagras_2018,Zadeh_1975}  have been used to model real-life processes in a wide range of applications such as robotics \cite{Hagras2004}, control systems \cite{Tseng2001}, weather prediction \cite{Zuoyong1998}, and stock market \cite{Mendel_2017_ch1}. The widespread adoption of XAI systems for modelling real-life processes is because of XAI systems' capability to integrate uncertainty in the input data, and subsequently improve the process' model and performance. In addition, the use of conceptual labels (CoLs) and prepositions by XAI systems pave way for human-understandable models for describing complex real-life processes. 
%The XAI systems are based on fuzzy logic systems \cite{Zadeh_1988}. 

The FLS based XAI systems handle uncertainty in the input data using fuzzy sets that convert crisp numbers (viz. uncertain observations) to CoLs characterised with membership values \cite{Zadeh_1975,Zadeh_1965}. The fuzzy sets are defined by membership functions (MFs) and represent a given CoL. The membership value is usually in the range [0,1] and is a soft measure of the degree of association the associated fuzzy set has for a given crisp measurement to belong to the CoL represented by the fuzzy set \cite{Zadeh_1965}. For example, an XAI system modelling the heights of people in a community using type-1 fuzzy sets may represent height using CoLs of \emph{Tall, Medium,} and \emph{Short}. The MF associated with each CoL's MF will assign a crisp number for the height of a person with a membership grade; for example, a height of 6ft may get assigned membership grades of \({0.8, 0.5, 0.1}\) to represent CoLs of \emph{Tall, Medium,} and \emph{Short} respectively. 

\textcolor{blue}{In general, fuzzy sets can model uncertainty in the feature domain at different levels: Type 1 (T1), interval type-2 (IT2), and general type-2 (GT2) fuzzy sets; illustrated in Fig. \ref{Fig:Types_of_Fuzzy_Sets}. Despite the variability in the extent for uncertainty modelling amongst the types of fuzzy sets, all fuzzy sets are modelling uncertainty from a single time snapshot of the feature domain. More specifically, fuzzy sets do not integrate associated temporal information in their membership grade calculation. This is a critical limitation of the fuzzy sets since most real-life systems are time-variant, i.e., their behaviour changes with time. To model time-dependent real-life systems more effectively, in this work, \textbf{we present the theory of a new \emph{Temporal Type-2 Fuzzy Set (TT2FS)} based approach for time-dependent XAI (TXAI)}}.

\begin{figure*}[tbp]
\begin{minipage}[b]{0.3\linewidth}
\subfloat[`Cold' membership function in temperature domain.]{\scalebox{.9}{%\begin{figure*}
%\centering
\resizebox{8cm}{5cm}{
\begin{tikzpicture}
\begin{axis}[no markers, domain=0:25, samples=100,
axis lines*=left, xlabel=Temperature \textdegree C, ylabel=Membership Degree \(\mu\),
 ymin =0, ymax= 1.1,
ytick = {0, 0.2, 0.4, 0.6, 0.8, 1.0},
xtick= {0, 5, 10, 15, 20, 25}, 
enlargelimits=false, clip=false, axis on top,
grid = major, xmin=0, xmax=25,
every axis plot/.append style={ultra thick}]
\addplot [name path = ColdU1, mark=none, persianblue!60, ultra thick] coordinates {(0,1) (8,1)};
%\addplot [name path = ColdL1, mark=none] coordinates {(0,.72) (8,.72)};
\addplot [name path = ColdU2, domain=8:25, persianblue!60] {gauss(8, 4, 10)} ;
%\addplot [name path = ColdL2, domain=8:20] {gauss(8, 4, 7.2)} ;
%\addplot [persianblue!60, opacity =0.4] fill between[of = ColdU1 and ColdL1, soft clip = {domain=0:8}];
%\addplot [persianblue!60, opacity =0.4] fill between[of =  ColdU2 and ColdL2, soft clip = {domain=8:20}];
\draw [dash pattern={on 5pt off 3pt on 1pt off 3pt}, draw =black, line width=.5mm] (15,0) -- (15,.2);
\draw [dash pattern={on 5pt off 3pt on 1pt off 3pt}, draw =black, line width=.5mm] (0,0.2) -- (15,.2);
% \node[coordinate, label={\small\textcolor{persianblue!80}{COLD}}] at (axis cs: 5, 1.0){};
\end{axis}
%\draw (current bounding box.south east) rectangle (current bounding box.north west);
\end{tikzpicture}
}
%\caption{Interval Type 2 Fuzzy Sets for event definition of Very Cold, Cold, Moderate, Hot, and Very Hot with respect to temperature values in \textdegree C.}
%\label{Fig:Universe_X}
%\end{figure*}
}} %\label{Fig:Cold_temp} 
\end{minipage} \hspace{2cm}
\begin{minipage}[]{0.7\linewidth}
\subfloat[`Cold' membership function in time domain.]{\scalebox{.7}{%----------------------------------------------------------
% \definecolor{mahogany}{rgb}{0.75, 0.25, 0.0}
% \definecolor{pgreen}{rgb}{0.0, 0.4, 0.0}
% \definecolor{persianblue}{rgb}{0.11, 0.22, 0.73}
% \definecolor{darkcyan}{rgb}{0.0, 0.55, 0.55}
% \definecolor{cadmiumred}{rgb}{0.89, 0.0, 0.13}
% \definecolor{amber}{rgb}{1.0, 0.75, 0.0}
%------------------------------------------------------------------------------------
% \pgfmathdeclarefunction{gauss}{3}{\pgfmathparse{(1*#3)/(#2*sqrt(2*pi))*exp(-((x-#1)^2)/(2*#2^2))}%
% }
%\begin{figure*}
%\centering
\adjustbox{valign=t, trim = 0pt 0pt 0pt 165pt}{
\begin{tikzpicture}
\pgfplotsset{every axis legend/.append style={
at={(0.5,1.03)},
anchor=south}}
\begin{axis}[no markers, domain=0:20, samples=100,
axis lines*=left, xlabel=Months of a year, ylabel=Membership Degree \(\mu\),
width = 15cm, height = 5cm, ymin =0, ymax= 1.1,
ytick = {0, 0.2, 0.4, 0.6, 0.8,1.0},
xtick={0, 4, 8, 12, 16, 20, 24, 28, 32,36, 40, 44, 48}, 
xticklabels={Dec, Jan, Feb, Mar, Apr, May, Jun, Jul, Aug, Sep, Oct, Nov, Dec},
x tick label style={rotate=90},
enlargelimits=false, clip=false, axis on top,
grid = major,
every axis plot/.append style={ultra thick},
legend columns=-1, 
legend style = {at={(.5,1.3)},anchor=south, draw=none},/tikz/every even column/.append style={column sep=0.5cm}]
\addplot [name path = Winter, mark=none, persianblue!60,ultra thick] coordinates {(0,1) (8,1)};
\addlegendentry{$Winter$}
%\addplot [name path = WinterL1, mark=none] coordinates {(0,.72) (4,.72)};
\addplot [name path = Spring, domain=8:20, pink,ultra thick] {gauss(8, 4, 10)} ;
\addlegendentry{$Spring$}
%\addplot [name path = WinterL2, domain=4:16] {gauss(4, 4, 7.2)} ;
%\addplot [persianblue!40, opacity =0.4] fill between[of = WinterU1 and WinterL1, soft clip = {domain=0:4} ];
%\addplot [persianblue!40, opacity =0.4] fill between[of = WinterU2 and WinterL2, soft clip = {domain=4:16} ];
%------------------------------------------------------------------------------------
\addplot [name path = Summer, domain=20:36, green!60!black,ultra thick] {gauss(8, 4, 10)} ;
\addlegendentry{$Summer$}
\addplot [name path = Autumn, domain=36:48, orange!60, ultra thick] {gauss(48, 4, 10)} ;
%\addplot [name path = ColdL1, domain=36:48] {gauss(48, 4, 7.2)} ;
%\addplot [persianblue!40, opacity =0.4] fill between[of = ColdU1 and ColdL1, soft clip = {domain=36:48} ];
\addlegendentry{$Autumn$}
%------------------------------------------------------------------------------------
\draw [dash pattern={on 5pt off 3pt on 1pt off 3pt}, draw =black, line width=.5mm] (10,0) -- (10,.85);
\draw [dash pattern={on 5pt off 3pt on 1pt off 3pt}, draw =black, line width=.5mm] (0,0.85) -- (10,.85);
\draw [dash pattern={on 5pt off 3pt on 1pt off 3pt}, draw =black, line width=.5mm] (14,0) -- (14,.3);
\draw [dash pattern={on 5pt off 3pt on 1pt off 3pt}, draw =black, line width=.5mm] (0,0.3) -- (14,.3);
% \node[coordinate, label={\small\textcolor{persianblue!80}{COLD}}] at (axis cs: 3, 1.0){};
\end{axis}
%\draw (current bounding box.south east) rectangle (current bounding box.north west);
\end{tikzpicture}}
%\caption{Interval Type 2 Fuzzy Sets for event definition of Very Cold, Cold, Moderate, Hot, and Very Hot with respect to the month of the year (time).}
%\label{Fig:Universe_T}
%\end{figure*}
}} %\label{Fig:Cold_time}
\end{minipage}
% \begin{minipage}[b]{0.45\linewidth}
%     \centering
%     \subfloat[a]{\includegraphics[trim= 1cm 2cm 1cm 2cm, clip, scale=0.45]{Figures/Cold_and_Time_UpperandLower_Mamdani.pdf} \label{Fig:Cold_temp}}
% \end{minipage}
\caption{An illustrative type-1 (T1) membership function (MF) for the fuzzy concept of `Cold' in the (a) universe of temperature in \textdegree C and (b) in the universe of time: months of a year. In this case, the membership degree for experiencing `Cold' at 15 \textdegree C is \(\mu_{{Cold}_{temp}}(15\)\textdegree C\() = 0.2\). 
Likewise considering the prevalence of particular linguistic variable 'Cold', viz. the \emph{likelihood of observing} `Cold' can be different in February \(\mu_{{Cold}_{time}}(February) = 0.85\) than March  \(\mu_{{Cold}_{time}}(March) = 0.3\). In this regard, the additional information of time can credit the primary membership in feature-domain through a fuzzy relation.} %In this regard, the additional information of time can contextualise the information of the primary membership in time-domain through a fuzzy relation.}
\label{Fig:Cold}
\end{figure*}
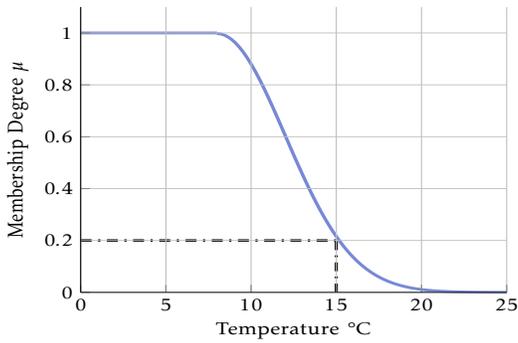
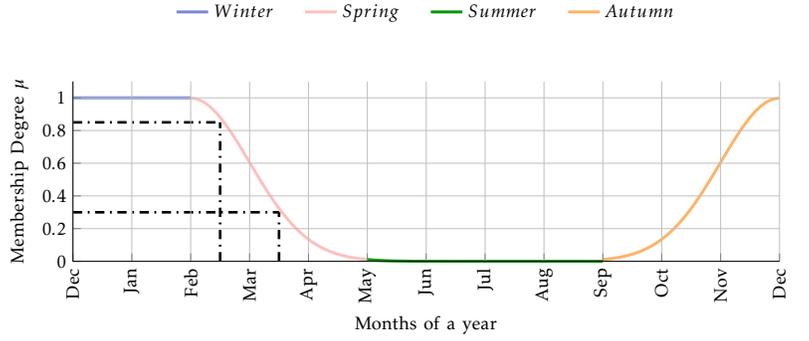

\textcolor{blue}{The prowess of TXAI system for incorporating time information for modelling time-variant processes is of paramount significance since the insights provided by a TXAI system can shed light on both spatial (feature domain) and temporal behaviour of the time-dependent process. More specifically, the TXAI is able to inform not only about the relation between input features but can also describe the impact of time on the evolution of the inter-relation of the features. As an example, let's consider a standard XAI system composed of a T1 fuzzy set for modelling thermal sensation `Cold' in the domain of values of temperature T \textdegree C as shown in Fig. \ref{Fig:Cold} (a), and a T1 fuzzy set for the time of occurrence of concept `Cold' during the months of a year as shown in Fig. \ref{Fig:Cold} (b). The notion is that the perception of `cold’ is mostly associated with the months of winter than in the months of spring. Hence, using the time information associated with a fuzzy concept (such as Cold in this case), a temperature can belong to the concept (Cold) differently according to a particular point in time (e.g., months of a year).} %  hence if we know that a given value of temperature of say 15\textdegree C (\(\mu_{Cold}(15\)\textdegree\(C)=0.2\)) is observed in the months of winter then it is more likely to be cold than if 15\textdegree C is observed in the months of spring i.e. \(\mu_{Cold}(15\)\textdegree\(C \;and\; mid \; February)\) should be a greater value than \(\mu_{Cold}(15\)\textdegree\(C \; and \; mid \; March)\). 

%----------------------------------------------------------
%----------------------------------------------------------
%---------------------------------------------------------- 
%--------------------------------------------------------
%Add neuroscience idea here
\textcolor{blue}{Crediting a fuzzy membership with its associated time information is particularly advantageous for the modelling of time-dependent noise-prone processes. Moreover, for dynamic processes, the ability to delineate its' (dynamic process) trajectories across time would inform the evolution of the temporal dynamics of the process.
To this end, our proposed TXAI system has been designed to integrate temporal information as well as able to outline the trajectories of a time-dependent process.} %(i.e., brain development trajectories for the case of functional brain development analysis).
To demonstrate the efficacy of TXAI system for time-dependent process modelling, in this work, an occupancy dataset is used \cite{Candanedo_2016_OccupancyDataset}. Using the values of temperature, light and carbon dioxide (CO\(_2\)), and the time the aforementioned measurements are taken, the TXAI system is used to make a prediction of whether or not the room is occupied. 

The rest of the paper is organised as follows: in Section II related works are outlined, Section III presents the TXAI system definition and operations, Section IV outlines the TXAI inference system with a numerical step-by-step example as well as the evolution of a TXAI model using temporal trajectories. An empirical study using TXAI system, as well as state-of-the-art systems (with varying levels of explainability) for performance comparison, on the aforementioned occupancy dataset \cite{Candanedo_2016_OccupancyDataset} is presented in Section V, with conclusion and future research in Section VI.

% The values of carbon dioxide (CO\(_2\)) in the room is estimated based on crisp values of the temperature and the light intensity in the room recorded at a given time of the day. 

%brain activity of surgeons whilst performing motor imagery tasks are modelled using TT2FSs. The brain signals of surgeons are acquired using functional near infrared spectroscopy signals (fNIRS). The motivation for using brain haemodynamic data, as a case study in this work, is because evoked brain signals vary significantly with respect to time. In this respect, brain signals which are characterised with uncertainty, owing to noise in measurements and inter-subject variability \cite{Gao_2014}, and also vary significantly with respect to time \cite{Meek_2002} makes them (brain signals) a viable case study to demonstrate the efficacy of TT2FSs to model complex, dynamic real-life systems.   

%---------------------------------------------------------------------
\section{Related Works}
%---------------------------------------------------------------------
Fuzzy sets %,which form the basis of XAI systems, 
have enabled explainable models of complex real-life processes which prove too ill-defined for closed form mathematical analysis. In this regard, although uncertainty in complex processes could be handled by fuzzy sets, the time-variant characteristics of complex processes have not been integrated into the modelling by standard XAI systems based on state-of-the-art fuzzy sets. 

There have been few notable attempts in the literature to model time in the MFs. The work by Garibaldi \emph{et al.} \cite{Garibaldi_2008} on \emph{non-stationary} fuzzy sets proposed that variation within a MF can be incorporated by perturbing the parameters of the MF. Their work aims to develop non-deterministic fuzzy reason as a way to model the variability in fuzzy decision making to mimic the variability in expert opinions. \textcolor{blue}{The ability of \emph{non-stationary} fuzzy sets to integrate differing experts' opinions is a significant contribution since it allows for a more comprehensive model that takes into account all experts' opinions.} However, their work does not incorporate the variation within a fuzzy concept with respect to time, which is the aim of the present work, to represent the time-variant transformation of a same fuzzy linguistic variable. 

Similarly, the work by Kostikova \emph{et al.} \cite{Kostikova_2016} propose \emph{dynamic} fuzzy sets by extending the classical fuzzy set to include a time dimension for representing MF at different time points. They propose four different types of dynamic MFs depending on how many parameters are changed in the definition of the dynamic MF. \textcolor{blue}{They simulated their dynamic MFs by using differing expert
assessments on multilevel fuzzy description of a complex system}. However, the dynamic MF is essentially a set of functions determined at different time points with no bearing on the temporal variation in the fuzzy concept. 

In another work by Maeda \emph{et al.} \cite{Maeda_1996}, they propose \emph{dynamic fuzzy reason} to deal with the notion of \emph{time delay} between premise and consequent. An example of where a time delay between premise and consequent assumes critical importance is: `If it starts snowing, the traffic on road will increase about 30 minutes later'. They propose the use of fuzzy relations between a fuzzy concept and its fuzzy time interval to assign a credit degree to the concept. The temporal fuzzy reasoning provides a framework for modelling delay in fuzzy reasoning and the temporal dynamics of a fuzzy concept. \textcolor{blue}{In this work, we have built on the work of Maeda \emph{et al.} \cite{Maeda_1996} to credit the membership grade of a concept based on time.}

To the best of the authors' knowledge, there is no work in the literature on fuzzy sets that delineates the incorporation of time-based variation in a fuzzy concept to compute the membership grade for the crisp values of the fuzzy concept. In addition, no previous work has aimed at delineating the trajectories of a time-variant process with respect to time. To this end, in this work, we propose TXAI systems that can integrate information from both the feature domain and time domain. More details on the proposed TXAI are outlined in Section \ref{sec:TXAI}.
% Another notion to address the issue of temporal information is \textcolor{blue}{there is also some thoughts on multi-variable membership functions 6.10 Mendel Book pp 300}

%---------------------------------------------------------------------
%---------------------------------------------------------------------
\section{Time-dependent Explainable Artificial Intelligence (TXAI) Systems} \label{sec:TXAI}
%---------------------------------------------------------------------
%---------------------------------------------------------------------
In this section, we present the TXAI system based on TT2FS (temporal type-2 fuzzy sets) that incorporate information from not only the uncertainty in the input domain of the fuzzy linguistic term, but also from its time of occurrence. In particular, the information from the time of occurrence is integrated into the membership grade of the TT2FS using fuzzy relations such that it (the membership grade of the TT2FS) varies with respect to time (time-dependent).

In the next section, we present the most common fuzzy relations and outline how they can be used for implementing TT2FS.
%----------------------------------------------------------
%----------------------------------------------------------
\subsection{Fuzzy relations between fuzzy linguistic variables and time related measures} \label{sec:fuzzyrelations}
%----------------------------------------------------------
%----------------------------------------------------------
In this work, fuzzy relations are used to interrelate the information with respect to the degree of truth of a determined linguistic term or CoL, $A$, within the domain X, and time, T, to form TT2FSs such that the likelihood of occurrence of A in \( x\in  X\), i.e. the primary membership grade \(\mu_A(x)\), is credited by a measure that is dependent on time such as frequency.
The application of fuzzy relation, for constructing TT2FSs, is motivated by the work on \emph{dynamic fuzzy reasoning models} in \cite{Maeda_1996}. They outline fuzzy relations that can be used to model time dependencies, as noted in Table \ref{Tab:Relations}.

Before reviewing the different relations that can be applied to construct a TT2FS, the conditions that need to be fulfilled by the associated temporal MF (TMF) are listed below:
\begin{enumerate}[label=(\roman*)]
    \item The TMF should be continuous.
    \item The TMF should be convex.
    \item The range of the TMF \(\subseteq[0,1]\).
    \item The TMF should reflect in the value of membership grade the intrinsic magnitudes of membership grade in feature domain and in frequency of occurrence domain, i.e., they should be directly proportional. For example, if \(\mu_A(x)\) is high and the time representation is also high then the result from the relation between them should also be high and vice versa. 
\end{enumerate}

\begin{table} [t]
\caption{Fuzzy relations between the universe of concept X and time domain T. }
\label{Tab:Relations}
\scalebox{1}{
\begin{tabular}{l l}
     \rowcolor{lightgray!20} 
     Name           & Definition of the relation\\
     Godel          &  \({R_{G}}(t,x)= 
   \begin{cases}
    1           & \text{if $\mu_{T_A}(t) \leq \mu_{A}(x)$} \\
    \mu_{A}(x)  & \text{if $\mu_{T_A}(t) > \mu_{A}(x)$} \\
  \end{cases}\) \\
   \rowcolor{lightgray!20} 
   Lukasiewicz    &  \(R_{L}(t,x)= 1 \wedge (1 - \mu_{T_A} (t) + \mu_{A} (x) )\) \\
   Gaines-Rescher & \(R_{GR}(t,x)= 
   \begin{cases}
    1           & \text{if $\mu_{T_A}(t) \leq \mu_{A}(x)$} \\
    0           & \text{if $\mu_{T_A}(t) > \mu_{A}(x)$} \\
  \end{cases}\)\\
  \rowcolor{lightgray!20}  
     Mamdani        & \(R_{M}(t,x) = \mu_{T_A} (t) \wedge \mu_{A} (x)\) \\
\end{tabular}}
\end{table}

%----------------------------------------------------------
%----------------------------------------------------------
%----------------------------------------------------------
%trim={<left> <lower> <right> <upper>}
\begin{figure*}
    \centering
    \includegraphics[scale=.4, trim = 3cm 0cm 3.8cm 0cm, clip]{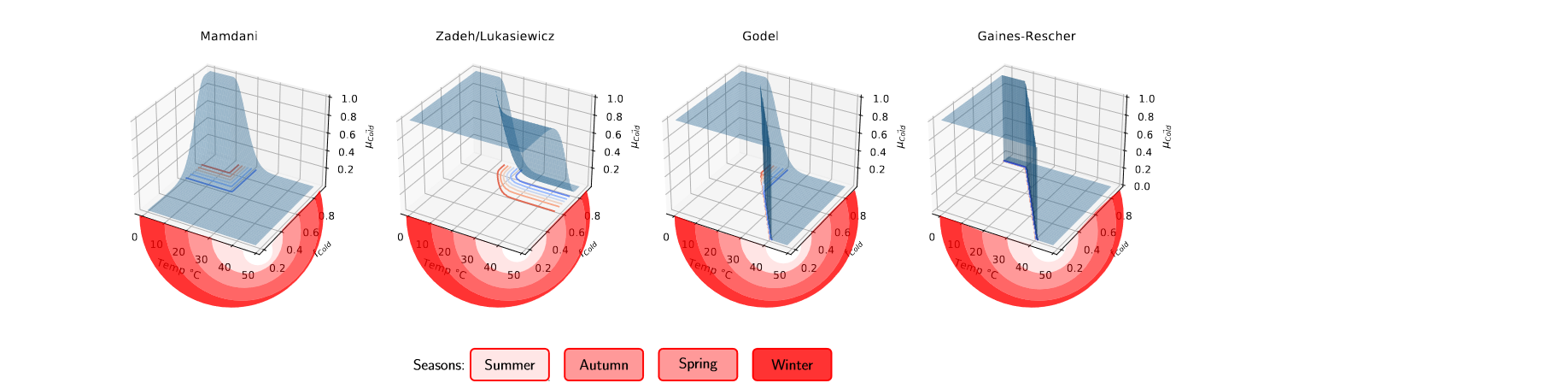}
    \caption{A comparison of TT2FSs for the conceptual label (CoL) `Cold' for feature thermal concept  constructed with the most commonly used fuzzy relations namely Mamdani, Zadeh/Lukasiewicz, Godel, and Gaines-Rescher, see Table \ref{Tab:Relations} for their respective definitions. In these illustrative plots, the feature domain i.e. temperature in \textdegree C is plotted on the x-axis, with conditional distribution, $f_{Cold}$ on y-axis, and the time is plotted on the axis connecting the x- and y- axis, i.e. the arc axis, with the 4 time intervals representing the typical seasons in a year. The z-axis has the values of temporal membership function (TMF), $\mu_{\vec{Cold}}(x, t, f_{Cold})$.}
    \label{Fig:Comp_Relations}
\end{figure*}
%----------------------------------------------------------
%----------------------------------------------------------

\noindent \textcolor{blue}{An illustrative comparison of the TT2FSs formed for the CoL `Cold' of feature thermal concept using the fuzzy relations listed in Table \ref{Tab:Relations} is shown in Fig. \ref{Fig:Comp_Relations}. The fuzzy relations are applied on hypothetical  primary membership function of `Cold' in feature domain (temperature) and time domain (months of a year). As can be seen in Fig. \ref{Fig:Comp_Relations}, the different fuzzy relations are encapsulating distinct inter-dependencies between time and feature domain. All relations meet the criteria (i) - (iii) listed above however, only the Mamdani relation meets the criterion (iv) as well since it gives credit to \(\mu_{Cold}\) based on the variable frequency of occurrence of `Cold' as observed in different months of the year. Hence, in this work, the Mamdani relation is used to construct the TT2FSs.}

% Based on the above desired characteristics of a temporal relation, we propose the following temporal relation which is  motivated from the Mamdani relation. 
% \begin{equation*}\label{Eq:Implication_Operator_Proposed}
%     \mu_{{T_E}\otimes E} =  
%   \begin{cases}
%     \mu_{T_E} (t) \wedge \mu_{E} (x)            & \text{if $\mu_{T_E}(t) \leq \mu_{E}(x)$} \\
%     \mu_{E} (x)           & \text{if $\mu_{T_E}(t) > \mu_{E}(x)$} \\
%   \end{cases}
% \end{equation*}

% To illustrate this using interval type 2 fuzzy sets (IT2FS), Fig. \ref{Fig:Universe_X} shows the conceptual event with respect to temperature, and the conceptual event with respect to month of the year.
\subsection{Conditional relative frequency distribution of a fuzzy linguistic term}
%As a measure relative to time we use use the frequency domain \(F\). 
In our TT2FS we employ a measure of conditional relative frequency between time and the occurrence of a linguistic term. We denote as $A$ an instance of a linguistic term from a set of conceptual labels (also called words of the universe of discourse), $CoLs := [CoL_{1}, CoL_{2}, ..., CoL_{J}]$ of a specific linguistic variable or input.

\begin{definition}[Discrete conditional relative frequency with respect to time] The discretized conditional relative frequency is defined as the likelihood of observing a linguistic term A %, given a fuzzy linguistic term, 
based on its membership grade, across time. This is denoted as \(g_A (t_n, \mu_A(x))\) with time \(t\) discretised over \(N\) time points (\(t_n\)) such as \(t_n \; \in \; [t_1,..., t_N]\), and is given by:
\begin{equation}\label{eq:FoCC}
    g_A (t_n, \mu_A(x)) = \frac{\sum\limits_{x \in X,t_n} \delta_{nj}}{\max\limits_{[t_1,..., t_N]}\left(\sum\limits_{x \in X,t_n} \delta_{nj}\right)}
\end{equation}
\(\delta_{nj}\) is a Kronecker delta function \cite{Kozen_2007_Kronecker} (e.g. $\delta_{ab}$ = 0 if a $\neq$ b, $\delta_{ab}$ = 1 if a=b) that takes the value of 1 when the following condition applies, $\exists\; {argmax}_j(\mu_{CoL_j}(x^{t_n})): Col_j = A$, $\forall j \in [1,...,J]$, and 0 otherwise. Note \(x^{t_n}\) is a realisation of \(x\) at time \(t_n\).
\end{definition}

The numerator in (\ref{eq:FoCC}) finds the count of occurrences of a given \(A\) for a determined time point \(t_n\) across all data instances, whereas the denominator is finding the maximum value of the count of occurrences of \(A\) across all \(N\) time points and all data instances. The resultant discrete conditional relative frequency \(g_A (t_n, \mu_A(x))\) is interpolated to form a conditional distribution \( f_A(t,\mu_{A}(x))\). For the sake of notational simplicity, we denote the later distribution as $f_A$ and the discrete conditional relative frequency as $g_A$ from here onwards.

Let us assume that the linguistic variable is thermal sensation defined on the input domain (\(x \in X\)) of temperature in \textdegree C and the associated CoLs be: [Cold, Comfortable, Hot]. For a given crisp input of temperature such as \(15\)\textdegree C, the associated primary membership grade for all three CoLs of Cold, Comfortable, and Hot be $\mu_{Cold}(15\)\,\textdegree\(C) = [0.4],\,\mu_{comf.}(15\)\textdegree\(C) = [0.3],\, \mu_{hot}(15\)\textdegree\(C) = [0]$ respectively. In this illustrative case, the temperature of 15 \textdegree C has a maximum membership grade, amongst all CoLs, for \emph{Cold} and hence 15 \textdegree C is assigned with the CoL of \emph{Cold}. Referring back to (\ref{eq:FoCC}), for computing the conditional relative frequency for \emph{Cold} the numerator is going to sum all the data instances where the crisp inputs are assigned with \emph{Cold} for a given time point \(t_n\) such as a particular month of a year. The denominator finds the mode of occurrence of \emph{Cold} across all months. The result of the division will scale the \(g_{Cold}\) values to [0,1]. 

An illustration for calculating the \(g_{Cold}\) values using  (\ref{eq:FoCC}), with a total of 12 time points as the months of a year %and the associated 4 time intervals can be the seasons in a year such as Winter, Spring, Summer, and Autumn, 
is shown in Fig. \ref{Fig:TemporalFuzzifier} (b) with continuous values of \(f_{Cold}\), found using interpolation of \(g_{Cold}\), plotted in Fig. \ref{Fig:TemporalFuzzifier} (c). Please note the associated time intervals, (as listed in the illustration in Fig. \ref{Fig:TemporalFuzzifier} are seasons in a year such as Winter, Spring, Summer, and Autumn), are for easing the computational complexity of the four-dimensional (4D) TT2FSs as will be explained later in section \ref{sec:Operators_TempFuzzySet} by taking time interval based slice of the TT2FS.

\subsection{Temporal Type-2 Fuzzy Sets (TT2FS)} \label{sec:TemporalFS}
%----------------------------------------------------------
%---------------------------------------------------------- \textcolor{blue}{}
%----------------------------------------------------------
In this section, a formal definition of temporal type-2 fuzzy sets (TT2FS) is presented. TT2FS are 4D as they incorporate information from the input domain (\(X\)), time domain (\(T\)), frequency of occurrence domain (\(F\)) and are characterised by a temporal membership function (TMF).

\noindent The computation of TMF, hereby termed as \emph{temporal fuzzification}, involves two stages: 1) fuzzification of crisp input values of \(A\) from feature domain \(X\) to form T1 \(\mu_A(x)\), as undertaken in standard T1 fuzzy sets; and 2) computation of the conditional distribution of \(A\), \(f_A\). The temporal fuzzification is illustrated in Fig. \ref{Fig:TemporalFuzzifier} (a) and defined next.

%----------------------------------------------------------
%----------------------------------------------------------
\begin{definition}[Temporal membership function]
The temporal membership function (TMF) can be defined as
\renewcommand\qedsymbol{$\blacksquare$}
\begin{equation} \label{eq:TMF}
\mu_{\vec{A}}(x, t, f_A) = \mu_A(x) \otimes f_A
\end{equation}
where \(\otimes\) is a relation operator, \(\mu_A(x)\) is the primary membership of \(A\) in feature domain credited by the conditional distribution of \(A\), denoted \(f_A\), using the Mamdani relation (outlined earlier in Sec \ref{sec:fuzzyrelations}). 
\\
%----------------------------------------------------------
%----------------------------------------------------------
\begin{theorem}
The TMF of A, constructed using Mamdani relation (\ref{eq:TMF}), \(\mu_{\vec{A}}(x, t, f_A)\) is \(\subseteq [0,1]\).
\end{theorem}
%----------------------------------------------------------
%----------------------------------------------------------
\begin{proof}
The range of \(\mu_{\vec{A}}(x, t, f_A)\) follows directly from the range of primary MF of A: \(\mu_A(x) \subseteq [0,1]\), and the conditional distribution of A: \(f_A \subseteq [0,1]\). Hence, by crediting \(\mu_A(x)\) with \(f_A\) using Mamdani relation (taking the min or product), it follows that the range of \(\mu_{\vec{A}}(x, t, f_A) \subseteq [0,1].\) 
\end{proof}
%----------------------------------------------------------
%----------------------------------------------------------
\begin{proposition}
If the primary membership of TMF is normal and the conditional distribution \(f\) is normal, according to (\ref{eq:FoCC}), then the resultant TMF membership after applying the Mamdani relation yields a normal temporal membership function, therefore we can imply that
\begin{equation}
\underset{x \in X}{sup} \: \mu_{\vec{A}}(x, t, f_A) = 1
\end{equation}
\end{proposition}
\begin{proof}
Given a $f_A \subseteq [0,1]$ and a \(\mu_A(x) \subseteq [0,1]\) both with $sup=1$, $\forall x \in X$ by deduction, $\exists x : f_A \times \mu_A(x) \lor min(f_A,\mu_A(x)) = 1$   
\end{proof}
%----------------------------------------------------------
\end{definition}
%----------------------------------------------------------
%----------------------------------------------------------
\noindent Next, we define the TT2FS  which are characterised by a TMF.
%----------------------------------------------------------
%----------------------------------------------------------
\begin{definition}[Temporal Type-2 Fuzzy Sets (TT2FS)]
%----------------------------------------------------------
A TT2FS \(\vec{A}\)
of the universe of discourse \(X \times T \times  F\) is characterised by a credited TMF \(\mu_{\vec{A}}(x,t, f_A): X \times T \times F \rightarrow [0,1]\) where X is the feature domain of A characterised by a T1 MF \(\mu_{A}(x)\), T is the time domain of A, \(F\) is the frequency of occurrence domain of A %characterised by a T1 MF \(f_{A}(t, \mu_A(x))\). 
characterised by conditional frequency distribution with respect to time \(f_{A}\). 
In mathematical set notation, \(\vec{A}\) can be written as (\ref{eq:TFset}):
%----------------------------------------------------------
\begin{align} \label{eq:TFset}
\begin{split}
\vec{A} = &\{ (x, t, f_A, \mu_{\vec{A}}(x, t, f_A)) \: |\\
& \forall x \in X, \forall t \in T, \forall \mu_A(x) \subseteq[0,1],\\
&\forall f_A \in F \subseteq[0,1]\}
\end{split}
\end{align}
%----------------------------------------------------------
where \(\mu_{\vec{A}}(x, t, f_A) \subseteq[0,1]\). Please note the conditional distribution, \(f_{A}\), is a continuous distribution interpolated from discrete conditional relative frequency, \(g_{A}\), and is defined mathematically earlier in (\ref{eq:FoCC}). \(\vec{A}\) can also be expressed as:
%----------------------------------------------------------
\begin{align} \label{eq:TFset_int}
    \vec{A} = \int_{x \in X} \int_{t \in T} \int_{f_A \in F} \mu_{\vec{A}}(x,t,f_A) /f_A/t/x
\end{align}
%----------------------------------------------------------
\end{definition}
%----------------------------------------------------------
%----------------------------------------------------------
% denoted by \(\vec{A}\), and the associated temporal MF (TMF), denoted by  \(\mu_{\vec{A}}\), for a feature A with \(J\) associated CoLs in the feature domain, i.e. \(CoL_{j, A} \; \in \;\{CoL_{1, A}, CoL_{2, A}, ..., CoL_{J, A}\} \) for \(j \; \in \; [1, ..., J]\), discretised over \(N\) time points (\(t_n\)) with \(Q\) associated number of time intervals (\(\Delta t_{q}\)), i.e.,  \(t_n \; \in \; [t_1, t_2, ..., t_{N-1},t_N]\) are the \(N\) discretised time points and \(\Delta t_q \; \in \; [\Delta t_1, ..., \Delta t_Q]\) are the total \(Q\) time intervals such that \(\Delta t_q = t_n - t_{n-1}\) where \(t_n\) and \(t_{n-1}\) represent two distinct time points, is presented. 
%-----------------------------------------------
%-----------------------------------------------
%-----------------------------------------------

where \(\int\int\int\) denotes the aggregation over all admissible values of \(x\), \(t\), and \(f_A\). The associated TMF, \(\mu_{\vec{A}}(x, t, f_A)) \subseteq [0,1]\), scales the \(\mu_A(x) \) based on its conditional distribution \(f_A\) as defined in (\ref{eq:TMF}).

% Theorem for constructing a temporal membership function 
% the theorem should prove that TMF [0,1]
% Also a proposition to 

%----------------------------------------------------------
%--------------------------------------------------------------
%--------------------------------------------------------------
% Define block styles
%--------------------------------------------------------------
%--------------------------------------------------------------
\tikzstyle{decision} = [diamond, draw, fill=blue!20, 
    text width=4.5em, text badly centered, node distance=3cm, inner sep=0pt]
\tikzstyle{block} = [rectangle, draw,  
    text width=2.8cm, text centered, rounded corners, minimum height=1.5cm, fill= blue!30]
\tikzstyle{line} = [draw, -latex']
\tikzstyle{cloud} = [draw, ellipse,fill=red!20, node distance=3cm,
    minimum height=2em]
\tikzstyle{dotted_block} = [draw=black!30!white, line width=1pt, dash pattern=on 1pt off 4pt on 6pt off 4pt, inner ysep=1mm,inner xsep=1mm, rectangle, rounded corners ]
\tikzstyle{myarrows} = [draw=black,solid,line width=.5mm, ->]
\tikzstyle{dashedline} = [draw=black,dashed,line width=.5mm]
%--------------------------------------------------------------
%--------------------------------------------------------------
%--------------------------------------------------------------
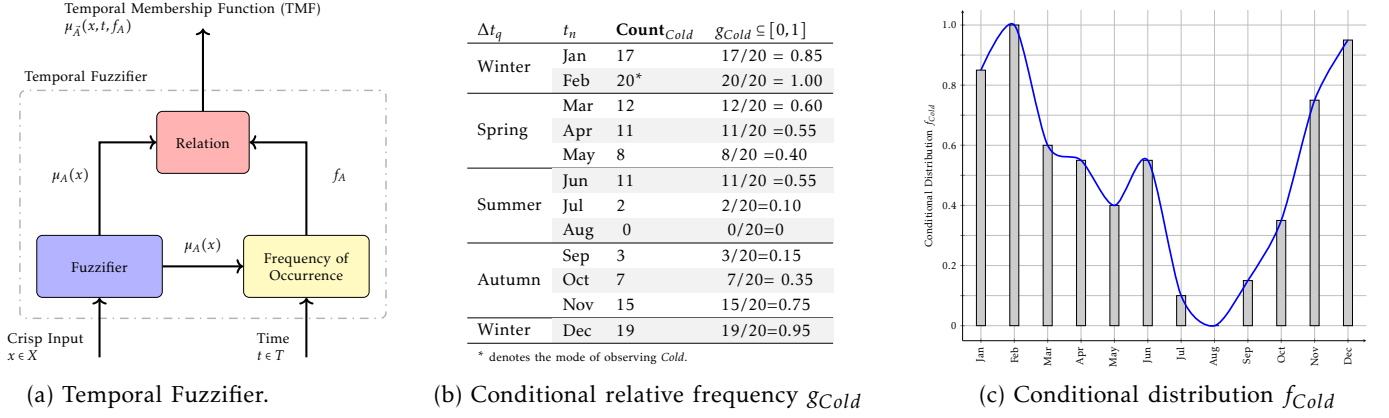
\begin{figure*}[!htbp]
\centering
\begin{tabular}{l l l}
     \scalebox{.55}{
\begin{tikzpicture}
    \node (fuzzifier) [block, fill= blue!30] {Fuzzifier};
    \node (FoO) [block, fill= yellow!30, right of = fuzzifier, node distance = 5cm] {Frequency of Occurrence};
    \node (relation) [block, above of = fuzzifier, text width = 2cm, fill= red!30, node distance = 3cm, xshift = 2.5cm] {Relation};
    \node (temporal_fuzzifier) [dotted_block, fit = (fuzzifier) (relation)(FoO), inner ysep=5mm,inner xsep=4mm] {};
    \node[above right] at (temporal_fuzzifier.north west){Temporal Fuzzifier};
    % Paths
    \draw[myarrows] (fuzzifier) -- node[xshift =0cm, yshift=.5cm]{\(\mu_A(x)\)}(FoO);
    \draw[myarrows] ++(0,-2.2) -- node[xshift = -1.3cm, yshift=-.5cm]{\begin{tabular}{l}
        Crisp Input  \\
        \(x \in X\) 
    \end{tabular}}(fuzzifier.south);
    \draw[myarrows] ++(5,-2.2) -- node[xshift =-.8cm, yshift=-.5cm]{\begin{tabular}{l}
        Time  \\
        \(t \in T\) \\
    \end{tabular}}(FoO.south);
    \draw[myarrows] (fuzzifier.north) |- node [xshift = -0.7cm, yshift=-0.8cm]{\(\mu_A(x)\)} (relation.west);
    \draw[myarrows] (FoO.north) |- node [xshift =.8cm, yshift=-0.8cm] {\(f_A\)} (relation.east);
    \draw[myarrows] (relation.north) -- node [yshift=1.25cm, xshift= -0.25cm]{\begin{tabular}{l}
         Temporal Membership Function (TMF) \\
        \( \mu_{\vec A}(x,t, f_A)\)\\
    \end{tabular}}++(0,2);
\end{tikzpicture}} 
& \hspace{.2cm}
%-----------------------------------------------------
%-----------------------------------------------------
\adjustbox{valign=t, trim = 0pt 0pt 0pt 125pt}{
\renewcommand{\arraystretch}{1.2}
\scalebox{.65}{
 \begin{tabular}{l l l l} 
        \textbf{\(\Delta t_q\)} &\textbf{\(t_n\)}& \textbf{Count$_{Cold}$}   & \textbf{\(g_{Cold} \subseteq [0,1]\)}\\ \hline
%         %---------------------------------
      \multirow{2}{*}{Winter}   & Jan & 17 & \FPeval{\result}{round(17/20,2)} 17/20 = \result  \\
%         %---------------------------------- \rowcolor{gray!10}
        &\cellcolor{gray!10}Feb & \cellcolor{gray!10}20\(^*\)& \cellcolor{gray!10}\FPeval{\result}{round(20/20,2)} 20/20 = \result\\ \hline
%       %----------------------------------
        \multirow{3}{*}{Spring}  & Mar & 12 &  \FPeval{\result}{round(12/20,2)} 12/20 = \result\\
%         %----------------------------------
      & \cellcolor{gray!10}Apr & \cellcolor{gray!10}11&  \cellcolor{gray!10}\FPeval{\result}{round(11/20,2)} 11/20 =\result \\  
%         %-------------------------
          & May & 8 &  \FPeval{\result}{round(8/20,2)} 8/20 =\result\\ \hline
%         %----------------------------
        \multirow{3}{*}{Summer} &\cellcolor{gray!10}Jun &\cellcolor{gray!10}11 & \cellcolor{gray!10} \FPeval{\result}{round(11/20,2)}11/20 =\result\\
%         %----------------------------
        &        Jul & 2&  \FPeval{\result}{round(2/20,2)} 2/20=\result\\
%         %----------------------------
        & \cellcolor{gray!10}Aug &\cellcolor{gray!10} 0 &\cellcolor{gray!10} \FPeval{\result}{round(17/20,2)} 0/20=0 \\  \hline
%         %----------------------------
         \multirow{3}{*}{Autumn} & Sep & 3&  \FPeval{\result}{round(3/20,2)} 3/20=\result \\
%         %----------------------------
        &  \cellcolor{gray!10}Oct & \cellcolor{gray!10}7 &  \cellcolor{gray!10} \FPeval{\result}{round(7/20,2)} 7/20= \result\\
%         %-------------------------------
        & Nov & 15 &  \FPeval{\result}{round(15/20,2)} 15/20=\result\\ \hline
%         %----------------------------------
         \multirow{1}{*}{Winter} &\cellcolor{gray!10}Dec & \cellcolor{gray!10}19 & \cellcolor{gray!10} \FPeval{\result}{round(19/20,2)}19/20=\result \\ \hline
%         %-----------------------
        \multicolumn{4}{l}{\scriptsize \(^*\) denotes the mode of observing \emph{Cold}.}\\
     \end{tabular}
     }}
     %--------------------------------------------------------
    %--------------------------------------------------------
     & \hspace{0.1cm}
     \pgfplotsset{ymin=0, ymax=20}
    \scalebox{.4}{\begin{tikzpicture}
\begin{axis}[
    %title=Title,
     axis lines=left, xtick=\empty,
     grid=both,
    width=15cm,
    enlargelimits=0.05,
    % legend style={at={(0.5,-0.15)},
    %   anchor=north,legend columns=-1},
      %ymin=0, ymax= 24,
    ylabel={Conditional Distribution \(f_{Cold}\)},
    bar width=3mm, y=5mm,
    symbolic x coords={Jan, Feb, Mar, Apr, May, Jun, Jul, Aug, Sep, Oct, Nov, Dec},
    xtick=data,
    x tick label style={rotate=90,anchor=east},
    ytick={0,2,4,6,8,10,12,14,16,18,20,22,24},
    yticklabels={0, , 0.2, ,0.4, , 0.6, , 0.8, , 1.0, ,}
    %nodes near coords align={vertical},
    ]
\addplot[ybar, fill=black!20] 
    coordinates {(Jan,17) (Feb,20) (Mar,12) (Apr,11) (May,8) (Jun,11) (Jul,2) (Aug,0) (Sep,3) (Oct,7)(Nov,15) (Dec,19) };
\addplot[draw=blue,ultra thick,smooth] 
    coordinates {(Jan,17) (Feb,20) (Mar,12) (Apr,11) (May,8) (Jun,11) (Jul,2) (Aug,0) (Sep,3) (Oct,7)(Nov,15) (Dec,19) };
\end{axis}
    \end{tikzpicture}}
     \\
     %--------------------------------------------------------
     %--------------------------------------------------------
    \hspace{0.5cm}\small (a) Temporal Fuzzifier. & \hspace{0.2cm}\small (b) Conditional relative frequency \(g_{Cold}\) & \small  \hspace{1cm} (c) Conditional distribution \(f_{Cold}\) %,\\
    %\hspace{.5cm} \(g_{Cold}\) & \hspace{1.25cm}\(f_{Cold}\) \\
\end{tabular}
%-----------------------------------------------------
%-----------------------------------------------------
\caption{(a) A schematic of temporal fuzzification for constructing temporal membership function (TMF). First, crisp values of input data from feature domain (i.e. \(x\in X\)) for a feature A are used to find primary membership function (MF) \(\mu_A(x) \). The values of \(\mu_A(x) \) associated with time \(t \in T\) are then transformed into a conditional distribution,\(f_A\), for each conceptual label (CoL) associated with A using discrete conditional relative frequency, \(g_{A}\) as outlined in (\ref{eq:FoCC}). The TMF for A, i.e. \( \mu_{\vec{A}}(x,t, f_A)\), is computed by applying a fuzzy relation (such as Mamdani relation) on \(\mu_A(x) \) and \(f_A(t, \mu_A(x))\). (b) A hypothetical calculation for conditional relative frequency of conceptual label \(Cold\) where A denotes the thermal sensation, i.e. \(g_{Cold}\) with respect to \(N=12\) discrete time points (\(t_n\)) i.e. the months and \(Q=4\) time intervals (\(\Delta t_q\)) representing the seasons in a year. The column Count$_{Cold}$ denotes the total number of times \(Cold\) was observed in the corresponding months i.e. the numerator in (\ref{eq:FoCC}). The mode for Count$_{Cold}$ is 20, and is observed in February, which becomes the denominator of (\ref{eq:FoCC}). (c) A bar plot of \(g_{Cold}\) for all individual discrete time points (months) with an interpolated continuous \(f_{Cold}\) superimposed in blue coloured solid line.}
\label{Fig:TemporalFuzzifier}
\end{figure*} 
\subsection{Operations on TT2FSs } \label{sec:Operators_TempFuzzySet}
%---------------------------------------------------------------------
%---------------------------------------------------------------------
%---------------------------------------------------------------------
In this section, the common operations for TT2FSs such as the union and intersection, as well as defuzzification are outlined. TT2FSs, on account of being 4D, are more computationally intense than GT2 fuzzy sets, which are three-dimensional (3D). A popular approach for minimising the computational demand of 3D GT2 fuzzy sets is to use z-slice based framework \cite{Wagner_2010}. Motivated from the effectiveness of z-slice based framework for simplifying the computations for GT2 fuzzy sets, in this work, the approach of taking time interval slice followed by z-slice (TS-ZS) is taken for performing operations on TT2FSs. The TS-ZS approach is explained in more detail as follows:

\begin{enumerate}[label=(\roman*)]
    \item TS: Time interval based slice to convert 4D TT2FSs into 3D. 
    % \begin{itemize}
    % \item 
    The 3D time interval based TT2FS is similar to 3D GT2 fuzzy set, with both sharing the feature domain on \(x-\)axis. On \(y-\)axis is the frequency of occurrence domain, for that time interval, for time interval based TT2FS, while for GT2 fuzzy sets, primary membership grade is on \(y-\)axis. And on \(z-\)axis is the temporal membership grade for time interval based TT2FS while for GT2 fuzzy set secondary membership grade is on \(z-\)axis.
    % \end{itemize}
    \item ZS: z-Slice based approach for the time interval based 3D TT2FS as utilised for GT2 fuzzy sets.
    % \begin{itemize}
        % \item
        The z-slices at specific z-levels render a given 3D fuzzy set to an equivalent IT2 fuzzy set with lower and upper primary membership grades. For the case of TS-ZS based TT2FSs, the primary membership grades are the conditional distribution values for that time interval at a given z-level. 
    % \end{itemize}
\end{enumerate}

\noindent In the following sections, a formal definition for the operations on TT2FSs is given with \(\vec{A}\) and \(\vec{B}\) denoting two TT2FSs characterised by TMFs \(\mu_{\vec{A}}(x,t,f_A)\) and \(\mu_{\vec{B}}(x,t,f_B)\) respectively as outlined in (\ref{eq:TMF_int}):

\begin{align} \label{eq:TMF_int}
    \vec{A} &= \int_{x \in X} \int_{t \in T} \int_{f \in F} \mu_{\vec{A}}(x,t,f_A) /f_A / t/x \nonumber \\
    \vec{B}&= \int_{x \in X} \int_{t \in T} \int_{f \in F} \mu_{\vec{B}}(x,t,f_B) /f_B / t/x
\end{align}

where \(X\) is the feature domain, \(T\) is the time domain, and \(F\) is the frequency of the occurrence domain. 
%----------------------------------------------
%--------- Algo for union and intersection -------------
%----------------------------------------------
%----------------------------------------------
\begin{algorithm}[!ht]
\setcounter{newAlgo}{1}
\SetAlgoLined
\KwResult{Resultant Temporal Membership Function (TMF) \(\mu_{\vec{A} \oslash \vec{B}}(x, t, f_{A \oslash B})\) where \(\oslash\) denotes the operation of either union or intersection.}
Let concepts A and B on feature domain X \textcolor{blue}{(input to the algorithm)} have TMFs denoted by $\mu_{\vec{A}}(x,t,f_A(t, \mu_A(x)))$ and $\mu_{\vec{B}}(x,t,f_B(t, \mu_B(x)))$ respectively with time intervals $\Delta t_q \in [\Delta t_1, ..., \Delta t_Q]$ and zslices discretised at $z_i \in [z_1, z_2, ..., z_I]$\;
For each time interval \(\Delta t_q\) the operation (union or intersection) on 3D time interval based TMF is computed independently by first taking the z-slices at $z_i \in [z_1, z_2, ..., z_I]$ which renders the 3D time interval based TMF into interval type 2 (IT2) TMFs\;
For each IT2 TMF, the operation is done as shown below in eq. (\ref{eq:algo_op})\;
 \For{$x \in X$}{
  \For{$z_i < z_I$}{
  \begin{equation} \label{eq:algo_op}
  \small \mu_{\vec{A} \oslash \vec{B}, \Delta t_q}(x, f_{\Delta t_q}) = \sum_{x} \sum_{f_{\Delta t_q} \in [\odot(l_A,l_B), \odot(u_A, u_B)]} z_i/f_{\Delta t_q}
  \end{equation}
 }
 }
 \small where the summation signs in eq. (\ref{eq:algo_op}) denotes the aggregation in set theoretic operation, \(l\) and \(u\) are the lower and upper conditional distribution values respectively of set \(\vec{A}\) and \(\vec{B}\) on z-slice $z_i$ and time interval $\Delta t_q$. For union operation, in  eq. (\ref{eq:algo_op}), the \(\odot\) denotes \(max\) and for intersection operation \(\odot\) denotes \( min\).
\caption{Union and Intersection Operations on TT2FSs}
\label{algo:Operation}
\end{algorithm}
%----------------------------------------------

%\input{Figures_texFiles_Used/Operators} 

%---------------------------------------------------------------------
%---------------------------------------------------------------------
\subsubsection{Union and Intersection Operations}
%---------------------------------------------------------------------
%---------------------------------------------------------------------
A general procedure for undertaking the union and intersection operations on the 4D TMFs is outlined in Algorithm \ref{algo:Operation}. The union of two TT2FSs \(\vec{A}\) and \(\vec{B}\) is a TT2FS defined as  \(\vec{A} \cup \vec{B}\) in  (\ref{Eq:Union_TFS}):

\begin{equation} \label{Eq:Union_TFS}
   \vec{A} \cup \vec{B}   = \int_{x \in X} \int_{t \in T} \int_{f \in F} \mu_{ \vec{A} \cup  \vec{B}}(x,t,f) / f /t / x 
\end{equation}

where \(\mu_{ \vec{A} \cup  \vec{B}}\) can be calculated by discretising the \(T\) domain, and 
taking z-slices on \(\mu_{ \vec{A} \cup \vec{B}, \Delta t_q}(x,t, f)\) values as outlined in (\ref{eq:algo_op}) of Algorithm \ref{algo:Operation}. In particular, for union operation, at time interval \(\Delta t_q\) (\ref{eq:algo_op}) takes the form of (\ref{Eq:Union_TFS_discrete}) when using the max t-conorm:

\begin{equation} \label{Eq:Union_TFS_discrete}
    \mu_{ \vec{A} \cup  \vec{B},\Delta t_q}(x,f_{\Delta t_q})  =\sum_{x} \sum_{f_{\Delta t_q} \in [max(l_A,l_B), max(u_A, u_B)]} z_i/f_{\Delta t_q}
\end{equation}

\noindent Likewise, the intersection of TT2FSs can be written as shown in (\ref{Eq:Int_TFS})

\begin{equation} \label{Eq:Int_TFS}
   \vec{A} \cap \vec{B}   = \int_{x \in X} \int_{t \in T} \int_{f \in F} \mu_{ \vec{A} \cap  \vec{B}}(x,t,f) / f /t / x 
\end{equation}

where \(\mu_{ \vec{A} \cap  \vec{B}}\) can be calculated by discretising the \(T\) domain, and taking z-slices on \(\mu_{ \vec{A} \cap \vec{B}, \Delta t_q}(x,t, f)\) values as outlined in   (\ref{eq:algo_op}) of Algorithm \ref{algo:Operation}. In particular, for intersection operation, at time interval \(\Delta t_q\)  (\ref{eq:algo_op}) takes the form of (\ref{Eq:Int_TFS_discrete}) when using the min t-norm. However, please note either product or min can be applied. 

\begin{equation} \label{Eq:Int_TFS_discrete}
    \mu_{ \vec{A} \cap  \vec{B},\Delta t_q}(x,f_{\Delta t_q})  =\sum_{x} \sum_{f_{\Delta t_q} \in [min(l_A,l_B), min(u_A, u_B)]} z_i/f_{ \Delta t_q}
\end{equation}
%---------------------------------------------------------------------
%---------------------------------------------------------------------

%---------------------------------------------------------------------
%---------------------------------------------------------------------
\subsubsection{Defuzzification}

In general, defuzzification converts a fuzzy set to an equivalent crisp number, and can be thought of as the inverse of fuzzification. For T1 fuzzy sets, defuzzification usually involves computing the centroid of the T1 fuzzy set \cite{Liang2000} to compute a representative crisp number, as shown in (\ref{Eq:Centroid}). 

%---------------------------------------------------------------------
%---------- Defuzzification Algorithm ---------------------
%---------------------------------------------------------------------
\setlength{\algomargin}{-5pt}
\begin{algorithm}[!btp]
\setcounter{newAlgo}{2}%
\SetAlgoLined
\KwResult{Crisp value for a given time interval, denoted by \(crisp_{\Delta t_q}\), where \(\Delta t_q\) is the \(qth\) time interval.}
Let feature A on feature domain X have temporal membership function (TMF) denoted by $\mu_{\vec{A}}(x,t,f_A(t, \mu_A(x)))$ with time intervals $\Delta t_q \in [\Delta t_1, ..., \Delta t_Q]$ and z-slices discretised at $z_i \in [z_1, z_2, ..., z_I]$\;
For each 3D time interval based TMF, the defuzzification can be done independently, by first taking the z-slices at $z_i \in [z_1, z_2, ..., z_I]$ which renders the 3D time interval based TMF into interval type 2 (IT2) MFs\;
The left and right centroid for each IT2 TMF at z-location \(z_i\), denoted by \(C_{z_i, \Delta t_q}\), can be computed using Karnik-Mendel (KM) method \cite{Chen_2021} to give [\(y_l, y_r\)] at that z-slice \(z_i\) and time interval \(\Delta  t_q\) as outlined in eq. (\ref{eq:defuzz})\;
 \For{\(z_i \leq z_I\)}{
  \begin{equation} \label{eq:defuzz}
        C_{z_i, \Delta  t_q} = [y_{l_{z_i, \Delta t_q}}, \;y_{r_{z_i, \Delta  t_q}}]
  \end{equation}
  %\(C_{Z,t_q}\).append(\(C_{z_i, t_q}\))
}
Defuzzifcation of the type reduced T1 fuzzy sets, using centroid average, to find  equivalent \(y_{l_{\Delta t_q}}\) and \(y_{r_{\Delta t_q}}\)\; 
\begin{equation} \label{eq:yl_crisp_value_algo}
    y_{l_{\Delta t_q}} = \frac{(z_1* y_{l_{z_1, \Delta t_q}}) + (z_2* y_{l_{z_2, \Delta t_q}}) +... + (z_I*y_{l_{z_I, \Delta t_q}})}{z_1 + z_2 + ... + z_I }
\end{equation}

\begin{equation} \label{eq:yr_crisp_value_algo}
    y_{r_{\Delta t_q}} = \frac{(z_1* y_{r_{z_1, \Delta t_q}}) + (z_2* y_{r_{z_2, \Delta t_q}}) +... + (z_I*y_{r_{z_I, \Delta t_q}})}{z_1 + z_2 + ... + z_I }
\end{equation}

A crisp value, \(crisp_{\Delta  t_q}\), can now be computed by applying Nie-Tan  method \cite{NieTan_2008} on \(y_{l_{\Delta t_q}}\) and \(y_{r_{\Delta t_q}}\).
%---------------------------------------------------------------------
\caption{Defuzzificaion of TT2FSs for a given time interval \(\Delta t_q\)}
 \label{algo:Defuzzification}
\end{algorithm}
%---------------------------------------------------------------------
%---------------------------------------------------------------------
\begin{equation} \label{Eq:Centroid}
    x^* = \frac{\sum_{b=1}^B x_b\mu (x_b)}{\sum_{b=1}^B \mu(x_b)}
\end{equation}

where \(x^*\) is the centroid of the T1 MF defined on the domain \(x \in X\). Here, the summation sign is used as in typical mathematical equations, i.e., for the case of the numerator, it is summing the product of \(x\) values and their corresponding membership values whereas for the denominator it is summing the membership values corresponding to all \(x_b\) values \(\forall b \in [1,..., B]\). 

For a 3D GT2 fuzzy set, defuzzification  usually involves three steps, outlined as follows:

\begin{enumerate}[label=(\roman*)]
\item Transforming a 3D GT2 fuzzy set to IT2 fuzzy sets by slicing the GT2 fuzzy set at given z-levels such as \(z_i \in [z_1, ..., z_I]\). 

\item Type reducing the z-level based IT2 fuzzy sets results in two T1 fuzzy sets using Karnik Mendel (KM) method \cite{Chen_2021}. The type-reduced T1 fuzzy sets are composed of the left and right centroids of the IT2 fuzzy sets. More specifically, the KM method requires iterative process to compute left and right centroids resulting in two T1 fuzzy sets: \([y_{l_{z_1}}, y_{l_{z_2}}, ..., y_{l_{z_I}}]\) and \([y_{r_{z_1}}, y_{r_{z_2}}, ..., y_{r_{z_I}}]\) where \(y_{l_{z_1}}\) is the left centroid at z-level 1 and \(y_{r_{z_1}}\) is the right centroid at z-level 1 and so on.
    %either:
    % \begin{itemize}
    % \item Karnik Mendel (KM) method \cite{Karnik_2001} which requires iterative process to compute left and right points usually denoted by \(y_l\) and \(y_r]\) 
    % \item Nie-Tan method which takes the average of the upper and lower MF of IT2 to compute the T1 MF \cite{NieTan_2008}. Nie-Tan methos is less computationally intense, in comparison to KM method, as the former is a closed form solution and hence does not require iterative process.
    % \end{itemize}

\item Defuzzifcation of the type reduced T1 fuzzy sets, using centroid average, to find  equivalent \(y_l\) and \(y_r\). 
\begin{equation} \label{eq:yl_crisp_value}
    y_l = \frac{(z_1* y_{l_{z_1}}) + (z_2* y_{l_{z_2}}) +... + (z_I*y_{l_{z_I}})}{z_1 + z_2 + ... + z_I }
\end{equation}

\begin{equation} \label{eq:yr_crisp_value}
    y_r = \frac{(z_1* y_{r_{z_1}}) + (z_2* y_{r_{z_2}}) +... + (z_I*y_{r_{z_I}})}{z_1 + z_2 + ... + z_I }
\end{equation}

\item The final type-reduced crisp value is found using the Nie-Tan method \cite{NieTan_2008} on \(y_l\) and \(y_r\).
\end{enumerate}

\noindent In this work, the defuzzification of 4D TT2FS also involves TS-ZS approach (explained earlier in section \ref{sec:Operators_TempFuzzySet}), i.e., taking the time interval based slice followed by z-slices. The time interval based TMF is 3D, and for each of the time interval (\(\Delta t_q\)) based TMF, z-slices at particular \(z_i\) levels renders them as IT2 fuzzy sets. The KM procedure \cite{Chen_2021} can be applied on IT2 fuzzy sets, at each z-level, to compute T1 fuzzy sets composed of \([y_{l_{z_i, \Delta t_q}}, y_{r_{z_i, \Delta t_q}}]\) as outlined in (\ref{eq:defuzz}). Using the centroid defuzzifier, the T1 fuzzy sets are defuzzified to give one equivalent \(y_l\) and \(y_r\), for that time interval, as outlined in (\ref{eq:yl_crisp_value_algo}) and (\ref{eq:yr_crisp_value_algo}). The Nie-Tan method \cite{NieTan_2008} is then applied to compute one crisp value for that time interval. The defuzzification of TT2FSs, for a given time interval, is summarised in Algorithm \ref{algo:Defuzzification}. The procedure outlined in Algorithm \ref{algo:Defuzzification} can be repeated for each time interval, i.e. \(\Delta t_q\) where \(q \in [1,..., Q]\), to obtain a crisp value for all time intervals.
%---------------------------------------------------------------------
\section{TXAI Inference System (TXAI-IS)}
\begin{figure*}[!hbtp]
\centering
\scalebox{0.7}{
\begin{tikzpicture} [node distance = 2.5cm]
    %---------------------------------------------
    \node (crisp_input) [block, fill= blue!30, text width = 2.5cm] {Crisp Input \(x \in X\) and \(t \in T\) };
    %---------------------------------------------
    \node (tempfuzzifier) [block, right of = crisp_input, text width = 2cm, fill= red!30, node distance = 3.5cm] { \begin{tabular}{l}
         Temporal  \\
         Fuzzifier 
    \end{tabular} };
    %---------------------------------------------
    \node (time_input) [block, fill= blue!30, above of = tempfuzzifier, text width = 2.5cm, node distance = 3.5cm] {Time intervals and CoL definitions};
    %---------------------------------------------
    \node (Temp_Rules) [block, right of = time_input, text width = 2cm, fill= green!30, node distance = 8cm] {Temporal Rules};
    %--------------------------------------------
    \node (Temp_Rules_timeInt) [block, below of = Temp_Rules, text width = 3cm, fill= green!20, node distance =  3.5cm] {\begin{tabular}{l}
         Temporal Rules \\
         at time interval 
    \end{tabular}};
    %--------------------------------------------
    % \node (Defuzzifier) [block, right of = Temp_Rules_timeInt, text width = 2cm, fill= yellow!30, node distance = 4.5cm] {Defuzzifier};
    %---------------------------------------------
    % \node (Type_Reducer) [block, right of = Temp_Rules_timeInt, text width = 2cm, fill= yellow!30, yshift = -2cm, xshift = 2cm] {Type Reducer};
    \node (zS_FLS) [block, right of = Temp_Rules_timeInt, text width = 3cm, fill= yellow!30, yshift = -5cm, xshift = 1.5cm] {zSlices based Fuzzy Logic System};
    %---------------------------------------------
    % \node (zslice) [block, below of = Type_Reducer, text width = 2cm, fill= yellow!30] {z-Slice};
    %---------------------------------------------
    \node (time_interval) [block, below of = tempfuzzifier, text width = 2cm, fill= cyan!20, node distance = 6.5cm, xshift = 3cm] {Time Interval \(\Delta t_q\) Slice};
    %---------------------------------------------
    \node (Inference) [block, below of = Temp_Rules_timeInt, text width = 2cm, fill= orange!30, node distance = 6.5cm] {Inference};
    %---------------------------------------------
    % \node (crisp_output_z) [block, right of = Defuzzifier, text width = 2cm, fill= brown!30, node distance = 3.5cm] {Crisp Output \(x^*\) for each z-level};
    
    %---------------------------------------------
    % \node (type_reduced) [block, right of = Type_Reducer, text width = 2cm, fill= magenta!30, xshift = 1cm] {Type Reduced T1-FS for all z-levels};
     \node (type_reduced) [block, above of = zS_FLS, text width = 3cm, fill= magenta!30, xshift = 0cm] {Type Reduced T1-FS };
    %---------------------------------------------
    % \node (crisp_output_tq) [block, right of = type_reduced, text width = 2cm, fill= brown!30, node distance = 3.5cm] {Crisp Output \(x^*\) for \(\Delta t_q\)};
    \node (crisp_output_tq) [block, above of = type_reduced, text width = 3cm, fill= brown!30, node distance = 2.5cm] {Centroid Calculation};
    % %---------------------------------------------
    % \node (repeat_zslice) [dotted_block, inner ysep=2mm, inner xsep=4mm, fit=(zslice)(Type_Reducer), label={[rotate=90, xshift=-2.5cm, yshift = 1.7cm]Repeat for each z-level}] {};
    %---------------------------------------------
    \node (repeat_time) [dotted_block, inner ysep=8mm, inner xsep=6mm, fit=(time_interval)(Inference)(Temp_Rules_timeInt)(type_reduced), label={[xshift=-4.7cm]Repeat for each time interval, \(\Delta t_q\)}] {};
    %---------------------------------------------
    % % Paths
    %---------------------------------------------
    \draw[myarrows] (crisp_input)--(tempfuzzifier.west);
    \draw[myarrows] (time_input)--(tempfuzzifier.north);% time_interval
    \draw[myarrows] (tempfuzzifier)|-(time_interval) node [pos=0.25,below, rotate=270] (TextNode) {\begin{tabular}{c}4D Temporal Fuzzy \\ Input Sets \end{tabular}};
    \draw [myarrows] (time_interval) -- (Inference) node [pos=0.45,below] (TextNode) {\begin{tabular}{l}3D Fuzzy \\ Input Sets \end{tabular}};
    \draw[myarrows] (Temp_Rules)--(Temp_Rules_timeInt);
    \draw[myarrows] (Temp_Rules_timeInt)--(Inference);
    \draw[myarrows] (Inference)-|(zS_FLS)node [pos=0.3, below] (TextNode) {\begin{tabular}{l} 3D Fuzzy\\ Output Sets \end{tabular}};
    %\draw[myarrows] (Type_Reducer)--(Defuzzifier)node [pos=0.3,left] (TextNode) {\begin{tabular}{l}T1-FS \end{tabular}};
    \draw[myarrows] (zS_FLS)--(type_reduced);
    \draw[myarrows] (crisp_output_tq)--++(5,0) node[above]{Crisp Output for \(\Delta t_q\), \(crisp_{\Delta t_q}\)};
    \draw[myarrows] (type_reduced)--(crisp_output_tq);
\end{tikzpicture}}
\caption{A general schematic representation delineating the interlinks between salient components of a time-dependent explainable artificial intelligence (TXAI) inference system (TXAI-IS).}
\label{Fig:TemporalInferenceMecahnism}
\end{figure*}
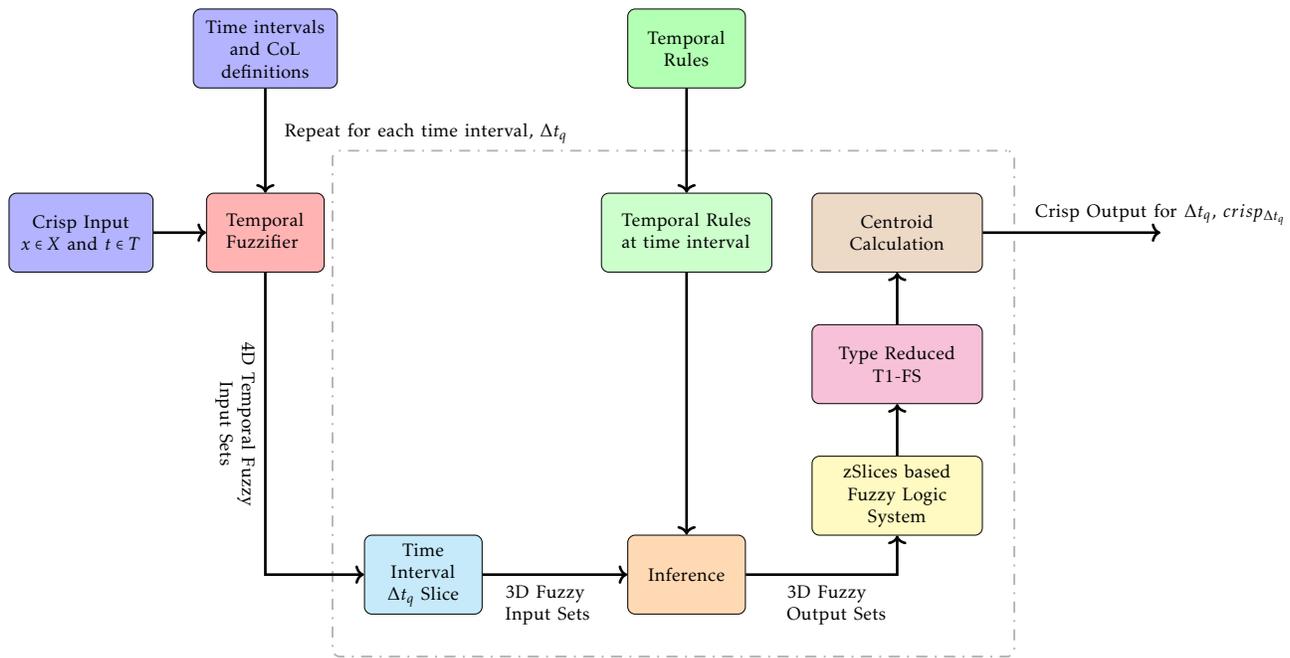

%---------------------------------------------------------------------
In this section, the TXAI inference system (TXAI-IS) for classification problems is outlined. %The inference mechanism are inherently different for classification and regression problems since in classification problem a label (or class) is predicted whereas in regression problem a numerical quantity is predicted. A common element for both regression and classification TXAI-IS is that the TT2FSs are split based on time intervals first and then based on the z slice level discretisation (TS-ZS), as described earlier in section \ref{sec:Operators_TempFuzzySet}. 
A general flowchart for the TXAI-IS is outlined in Fig. \ref{Fig:TemporalInferenceMecahnism}. The temporal fuzzifier constructs the 4D TT2FSs as outlined in Fig. \ref{Fig:TemporalFuzzifier} (a). To analyse a given dynamic process with respect to time, the TXAI-IS works for each time interval \(\Delta t_q\) where \(\Delta t_q \in [\Delta t_1, ...,\Delta t_Q ]\) independently. To this end, the 4D TT2FSs are first sliced based on the \(\Delta t_q\), and inference is made on time sliced 3D TT2FSs using the temporal rules for the same \(\Delta t_q\). Each time interval would entail a unique temporal rule base. The temporal rules can either be furnished by experts in the field or can be learnt from the input data using evolutionary algorithms such as genetic algorithm (GA) \cite{Mirjalili_2019}. 

\textcolor{blue}{In addition, the assumptions of the proposed TXAI system with TT2FSs include: 1) the input features and output are observable, 2) a relation between input features and output exists, and 3) the relation between input features and output varies with time. }

In the next subsections, the classification TXAI-IS is outlined in detail as the empirical study on which TXAI system is exemplified also undertakes a classification problem, i.e., occupancy dataset \cite{Candanedo_2016_OccupancyDataset} is analysed to determine whether or not a room is occupied. 

\subsection{Classification} 
%---------------------------------------------------------------------
%---------------------------------------------------------------------
For the classification problem, the TXAI-IS will predict one class or label for a given data instance for each time interval. The overall TXAI-IS for classification undertakes the following steps:

\begin{enumerate}[label=(\roman*)]

\item Compute the membership degree for the time interval based 3D TT2FSs. 
%\item Compute the degree of membership.
\begin{itemize}
\item The time interval based 3D TT2FSs are transformed into IT2 fuzzy sets by taking slices at predefined z-levels. The degree of membership at each z-level, such as \(z_i \in [z_1, ..., z_I]\) where \(I\) is the total number of z slices, for a given 3D TT2FS A is given as follows \cite{Wagner_2010}:
\begin{align}\label{eq:MemDeg}
    \tilde{A} = \{(x,u,z)|&\forall x \in X, \\ \nonumber
    & \forall u \in [\underline{\mu}_{\tilde{A}}(x),\overline{\mu}_{\tilde{A}}(x)]\subseteq[0,1]\}
\end{align}
where \(\mu_{\tilde{A}}\) is the membership degree of the IT2 fuzzy set \(\tilde{A}\) at the predefined \(z\) level. 
\end{itemize}
%----------------------------------------------------------------
%----------------------------------------------------------------
\item Compute the firing strength for each rule, at each z-level. 
\begin{itemize}
\item The \(\overline{upper}\) and \(\underline{lower}\) firing strength of a given rule \(p\), \(\overline{w}_p\) and \(\underline{w}_p\) respectively, is the degree of match between the rule \(p\) and the data instance \(x\). It is computed as:
\begin{align}\label{eq:SOA}
    \overline{w}_p (x^k)= & \prod_{k=1}^a \overline{\mu}_{\tilde{A}}(x^k) \nonumber \\
    \underline{w}_p (x^k) = & \prod_{k=1}^a \underline{\mu}_{\tilde{A}}(x^k)
\end{align}
where \(p\) is the rule number, \(a\) is the total number of antecedents in the rule \(p\) and \(x^{k}\) is an input ($k$) of the actual data instance to be classified. 
\end{itemize}
\item Compute the rule weight (RW) for each rule, at each z-level.  
\begin{itemize}
    \item The RW is a measure of a given rule's dominance and is computed as shown in (\ref{eq:RW}).
    \begin{align} \label{eq:RW}
        \overline{RW}_p & = \overline{c}_p \times \overline{s}_p \\ \nonumber
        \underline{RW}_p & = \underline{c}_p \times \underline{s}_p
    \end{align}
    where $c$ is the confidence of the rule $p$ and $s$ is the support of the $p$th rule.
    \item The confidence of a rule is a measure of the likelihood to correctly classify a given data instance. It is calculated as shown in eq. (\ref{eq:conf})
    \begin{align} \label{eq:conf}
        \overline{c}_p (Ants_p \Rightarrow Cons_p) &= \frac{\sum_{x\in (Ants_p \Rightarrow Cons_p)} \overline{w}_p (x)}{\sum_{p=1,x\in (Ants_p)}^P \overline{w}_p (x)}  \\ \nonumber
        \underline{c}_p (Ants_p \Rightarrow Cons_p) &= \frac{\sum_{x\in (Ants_p \Rightarrow Cons_p)} \underline{w}_p (x)}{\sum_{p=1,x\in (Ants_p)}^P \underline{w}_p (x)}
    \end{align}
    where \(Ants_p\) and \(Cons_p\) are the antecedents and consequent respectively of the rule \(p\). The numerator sums the firing strength of all the data instances that have the same antecedents and consequent as the rule \(p\). Whereas the denominator sums the firing strength of all the data instances that have the same antecedents as the rule \(p\) irrespective of the consequent- for all the rules \([1, ..., P]\), where \(P\) is the total number of rules. 
    \item The support of a rule is calculated as shown in (\ref{eq:supp})
    \begin{align} \label{eq:supp}
        \overline{s}_p (Ants_p \Rightarrow Cons_p)&= \frac{\sum_{x\in (Ants_p \Rightarrow Cons_p)} \overline{w}_p (x)}{P}  \\ \nonumber
        \underline{s}_p (Ants_p \Rightarrow Cons_p) &= \frac{\sum_{x\in (Ants_p \Rightarrow Cons_p)} \underline{w}_p (x)}{P} 
    \end{align}
    with \(P\) as the total number of rules.
\end{itemize}
%---------------------------------------------------------------------
%---------------------------------------------------------------------
\item Compute the association degree of each rule, with a given data instance, for each z-level.
\begin{itemize}
    \item The association degree of a rule \(p\) with a given data instance \(x\) is computed as shown in (\ref{eq:AssoDeg}):
    \begin{align} \label{eq:AssoDeg}
        \overline{h}_p = \overline{w}_p(x) \times \overline{RW}_p \\ \nonumber
        \underline{h}_p =\underline{w}_p(x) \times \underline{RW}_p
    \end{align}
\end{itemize}
%---------------------------------------------------------------------
%---------------------------------------------------------------------
\item Predict the label.
\begin{itemize}
    \item Find a value of the association degree, \(h\), for each rule by using Nie-Tan \cite{NieTan_2008} method on the \(\underline{h}\) and \(\overline{h}\) which are found using (\ref{eq:yl_crisp_value_algo}) and (\ref{eq:yr_crisp_value_algo}).
    \item The rule with the highest association degree, \(h\), predicts the label for the given data instance.
\end{itemize}
%---------------------------------------------------------------------
\item The steps outlined above (i)-(v) are repeated for each time interval to predict a label for all time intervals.
%---------------------------------------------------------------------
\end{enumerate}
%---------------------------------------------------------------------
%---------------------------------------------------------------------
%---------------------------------------------------------------------
%---------------------------------------------------------------------
%---------------------------------------------------------------------
\subsection{Numerical Step-wise Example}
\label{sec:NumEx}
%---------------------------------------------------------------------
%---------------------------------------------------------------------
In this section, a binary classification problem using TXAI-IS is exemplified using a hypothetical dataset with two input features, \emph{Feature1} and \emph{Feature2}, and one output. Let time intervals be defined over a day such as Morning, Daytime, and Evening with three CoLs associated with the inputs (\emph{Feature1} and \emph{Feature2}) be: [Low, Medium, High] and output labels be \emph{Output1} and \emph{Output2}. %Let the input data instance be: [Feature1 = 19.7, Feature2 = 4.3] for which the output label (\emph{Output1} or \emph{Output2}) will be predicted using TXAI-IS. 

First, TT2FSs for both inputs (\emph{Feature1} and \emph{Feature2}) are constructed using temporal fuzzifier, as outlined in Fig. \ref{Fig:TemporalFuzzifier}. Also, for each time interval, the rules will be different but the overall process to determine the output label is same. In the following steps, we exemplify how the output label is predicted for one time interval, in this example, Morning. 

Let the rules (R) outlining the relation between input features and output for Morning be as listed in (\ref{Eq:Class_Rules}). The corresponding lower and upper rule weights (RW) at each z-level are as listed in Table \ref{tab:Class_FI}. In the following steps i)- iv) we show how a corresponding label for \(Output\) is predicted using TXAI-IS for input values of \emph{Feature1} = 19.7 and \emph{Feature2} be = 4.3. In this example, the z-level is discretised at \(z_{0.2}\), \(z_{0.4}\), \(z_{0.6}\), \(z_{0.8}\), and \(z_{1.0}\).

\begin{align} \label{Eq:Class_Rules}
  R_1:& \text{ IF } \emph{Feature1} \text{ is } Low \text{ and } \emph{Feature2} \text{ is } Medium  \nonumber\\& \text{ THEN } \emph{Output} \text{ is } \emph{Output2}  \nonumber \\ %-----------------------------------------------------------
  R_2:& \text{ IF } \emph{Feature1}  \text{ is } Medium \text{ and } \emph{Feature2} \text{ is } Medium  \nonumber\\& \text{ THEN } \emph{Output} \text{ is } \emph{Output1} \nonumber \\
  %-----------------------------------------------------------
  R_3:& \text{ IF } \emph{Feature1}  \text{ is } High \text{ and } \emph{Feature2} \text{ is } High  \nonumber\\& \text{ THEN } \emph{Output} \text{ is } \emph{Output1}  \\ \nonumber
\end{align}

%---------------------------------------------
%---------- Classification Step by Step 
%---------------------------------------------
\begin{enumerate}[label=(\roman*)]
\item The degree of membership for each CoL of the inputs \emph{Feature1} and \emph{Feature2} is determined from the time interval (Morning) based 3D TMF. The membership degree is the value of the conditional distribution at a given input value and corresponding z-level as outlined in (\ref{eq:MemDeg}). Let the corresponding membership degrees for each CoL of the inputs \emph{Feature1} and \emph{Feature2} be as noted in Table \ref{tab:Class_MD_Ants}.
%---------------------------------------------------
%---------------- Membership Degree Antecedents -------------------
%---------------------------------------------------
%---------------------------------------------------
\begin{table}[h]
    \centering
    \caption{The hypothetical lower (L) and upper (U) degree of membership values of the conceptual labels (CoLs) of \emph{Feature1} and \emph{Feature2} for the time interval Morning for five z levels: \(z_{0.2}\), \(z_{0.4}\),  \(z_{0.6}\), \(z_{0.8}\), and \(z_{1.0}\) with input value of \emph{Feature1} = 19.7, and \emph{Feature2} = 4.3.}
    \begin{tabular}{l|l|l|l|l|l|l|l}
         \textbf{CoLs} & \textbf{CoLs} & &\textbf{\(z_{0.2}\)} & \textbf{\(z_{0.4}\)} &\textbf{\(z_{0.6}\)} & \textbf{\(z_{0.8}\)} & \textbf{\(z_{1.0}\)}   \\ \hline
         %----------------------------------------------------
          % Temp - Low
         \multirow{6}{*}{\emph{Feature1}} & \multirow{2}{*}{Low} &\cellcolor{gray!10}L &\cellcolor{gray!10}0.50& \cellcolor{gray!10}0.52&\cellcolor{gray!10}0.54 &\cellcolor{gray!10}0.52 &\cellcolor{gray!10}0.51\\
         &  & U & 0.61 & 0.63 & 0.64 & 0.61 & 0.60  \\ 
         %---------------------------------------------------
        % Temp - Medium
         & \multirow{2}{*}{Med.} &\cellcolor{gray!10}L &\cellcolor{gray!10}0.63 & \cellcolor{gray!10}0.63&\cellcolor{gray!10}0.65 &\cellcolor{gray!10}0.63 &\cellcolor{gray!10}0.61  \\
         &  & U & 0.77 & 0.78& 0.78& 0.77& 0.75 \\
         %---------------------------------------------------
        % Temp - High
         &  \multirow{2}{*}{High} &\cellcolor{gray!10}L &\cellcolor{gray!10}0.65 & \cellcolor{gray!10}0.64 &\cellcolor{gray!10}0.64 &\cellcolor{gray!10}0.63 &\cellcolor{gray!10}0.63 \\ 
         &  & U &0.69 &0.69 &0.68 &0.68 & 0.67 \\ \hline
         %---------------------------------------------------
         %---------------------------------------------------
         % Light - Low
          \multirow{6}{*}{\emph{Feature2}} & \multirow{2}{*}{Low} &\cellcolor{gray!10}L &\cellcolor{gray!10}0.31 & \cellcolor{gray!10}0.31&\cellcolor{gray!10}0.31 &\cellcolor{gray!10}0.31 &\cellcolor{gray!10}0.31\\
         &  & U &0.32 &0.32 & 0.32 &0.32 &0.32  \\ 
         %---------------------------------------------------
         % Light - Med
         & \multirow{2}{*}{Med.} &\cellcolor{gray!10}L &\cellcolor{gray!10}0.50 & \cellcolor{gray!10}0.55 &\cellcolor{gray!10}0.55 &\cellcolor{gray!10}0.54 &\cellcolor{gray!10}0.53  \\
         &  & U & 0.58 & 0.59 & 0.59 & 0.58 & 0.57  \\
         %---------------------------------------------------
          % Light - High
         &  \multirow{2}{*}{High}&\cellcolor{gray!10}L &\cellcolor{gray!10}0.40 & \cellcolor{gray!10}0.40&\cellcolor{gray!10}0.40 &\cellcolor{gray!10}0.42 &\cellcolor{gray!10}0.44\\ 
         &  & U & 0.43 & 0.43 & 0.46 & 0.46 & 0.49  \\ \hline
         %---------------------------------------------------
    \end{tabular}
    \label{tab:Class_MD_Ants}
\end{table}

\item The firing strength of each rule listed in (\ref{Eq:Class_Rules}) are found, using the membership degree in Table \ref{tab:Class_MD_Ants}, as outlined in (\ref{eq:SOA}) and listed in Table \ref{tab:Class_FI}. As an example, for \(R_1\) the lower firing strength at \(z = 0.6\), \(\underline{w}_{1_{z=0.6}}\), can be calculated as follows:
\begin{align}
    \underline{w}_{1_{z=0.6}} (x = [19.7, 4.3])&= \prod_{k=1}^2 \underline{\mu} (x^k) \nonumber \\
    & = 0.54 * 0.55 = 0.297
\end{align}

%---------------------------------------------------------
%---------- Firing Interval -------
%---------------------------------------------------------
\begin{table*}[t]
    \centering
    \caption{The lower and upper firing strengths, \(\underline{w}\) and \(\overline{w}\) respectively, for the hypothetical rules listed in (\ref{Eq:Class_Rules}) for time interval Morning. The rule weights (RW) at each z-level are also listed. }
    \label{tab:Class_FI}
    \renewcommand{\arraystretch}{1.2}
    \begin{tabular}{c| l| l | c| c| c| c | c| l| c| c| c| c| c}\hline
      \multirow{2}{*}{Rule}  & Firing & \multicolumn{5}{c}{z-level}  & \multirow{2}{*}{Consequent}& \multirow{1}{*}{Rule Weight}& \multicolumn{5}{c}{z-level} \\ 
      & Strength, \(w\)  &\(z_{0.2}\) & \(z_{0.4}\) & \(z_{0.6}\) & \(z_{0.8}\) & \(z_{1.0}\) & & RW &\(z_{0.2}\) & \(z_{0.4}\) & \(z_{0.6}\) & \(z_{0.8}\) & \(z_{1.0}\)\\\hline
%-------------------------------------------------------
% FS Lower : [0.25, 0.286, 0.297, 0.281, 0.27]
% FS Upper : [0.354, 0.372, 0.378, 0.354, 0.342]
% Lower: [0.31, 0.3, 0.3, 0.29, 0.27]
% Upper: [0.35, 0.34, 0.34, 0.31, 0.3]
  \multirow{2}{*}{\(R_1\)}  & Lower & 0.25&  0.286&  0.297&  0.281&  0.27 &\multirow{2}{*}{\emph{Output2}} & Lower & 0.31 & 0.30 & 0.30 & 0.29 & 0.27  \\
  & Upper & 0.354& 0.372& 0.378& 0.354& 0.342 & &Upper & 0.35 & 0.34 & 0.34 & 0.31 & 0.30\\ \hline
%-------------------------------------------------------
% FS Lower : [0.315, 0.347, 0.358, 0.34, 0.323]
% FS Upper :[0.447, 0.46, 0.46, 0.447, 0.427]
% Lower:[0.69, 0.69, 0.68, 0.66, 0.66]
% Upper: [0.73, 0.73, 0.72, 0.72, 0.72]
    \multirow{2}{*}{\(R_2\)} &Lower  & 0.315& 0.347& 0.358& 0.34& 0.323 & \multirow{2}{*}{\emph{Output1}}  & Lower & 0.69 & 0.69 & 0.68 & 0.66 & 0.66 \\
 & Upper & 0.447 & 0.46 & 0.46 & 0.447 & 0.427& & Upper & 0.73 & 0.73 & 0.72 & 0.72 & 0.72 \\ \hline
%-------------------------------------------------------
% FS Lower:[0.26, 0.256, 0.256, 0.265, 0.277]
% FS Upper:[0.297, 0.297, 0.313, 0.313, 0.328]
%RW Lower:  [0.22, 0.21, 0.21, 0.21, 0.21]
%RW upper: [0.24, 0.22, 0.22, 0.22, 0.22]
\multirow{2}{*}{\(R_3\)}  &Lower  &0.26& 0.256& 0.256& 0.265& 0.277 & \multirow{2}{*}{\emph{Output1}} & Lower& 0.22& 0.21& 0.21& 0.21& 0.21  \\ 
&Upper & 0.297& 0.297& 0.313& 0.313& 0.328  & & Upper & 0.24 & 0.22 & 0.22 & 0.22 & 0.22\\\hline
%-------------------------------------------------------
    \end{tabular}
\end{table*}

\item The association degree of each rule with the input data instance is determined, using the firing strength in Table \ref{tab:Class_FI}, as outlined in (\ref{eq:AssoDeg}). The upper and lower values of the association degree for the five z-levels are as listed in Table \ref{tab:Class_AD}. As an example, for \(R_2\) the upper association degree at \(z = 0.2\), \(\overline{h}_{2_{z=0.2}}\), can be calculated as follows:

\begin{align} 
    \overline{h}_{2_{z=0.2}} &= \overline{w}_{2_{z=0.2}}(x) \times \overline{RW}_{2_{z=0.2}} \\ \nonumber
    &=0.447 * 0.73 = 0.326
\end{align}

\item The consequent of the rule with the highest association degree with the input data instance becomes the predicted label for a given time interval. The crisp value for the association degree of each rule is found using (\ref{eq:yl_crisp_value_algo}) and (\ref{eq:yr_crisp_value_algo}). As an example, the crisp value of association degree for \(R_3\) is found as follows:
\begin{align}
    h_{3_{l}}& = \frac{0.2*(\underline{h}_{3_{0.2}})+ ... + 1.0*(\underline{h}_{3_{1.0}})}{0.2 + 0.4 + 0.6+0.8+1.0 } \nonumber \\
    & = \frac{0.2*0.057 + 0.4*0.054+...+ 1*0.058}{3} = 0.056\nonumber \\
 h_{3_{u}}& = \frac{0.2*(\overline{h}_{3_{0.2}})+ ... + 1.0*(\overline{h}_{3_{1.0}})}{0.2 + 0.4 + 0.6+0.8+1.0 } \nonumber \\
    & = \frac{0.2*0.071 + 0.4*0.065+...+ 1 *0.072}{3} = 0.0696\nonumber \\ h_{3_{crisp}}& = \frac{0.056+0.0696}{2}= \frac{0.1256}{2}=0.063
\end{align}

\noindent In this illustrative example, \(R_2\) has the highest association degree (tabulated in Table \ref{tab:Class_AD}) hence the predicted output for the input data instance (\emph{Feature1} = 19.7 and \emph{Feature2} be = 4.3) for time interval Morning is the consequent of \(R_2\), i.e., \emph{Output1}.
\end{enumerate}
%---------------------------------------------------------
%----------------- Classification: Association Degree -----------------------
%---------------------------------------------------------
\begin{table}[t]
\caption{The lower (L) and upper (U) association degrees, \(h\), for each of the three rules (R) listed in (\ref{Eq:Class_Rules}) with input data instance: Feature 1= 19.7, Feature 2 = 4.3. The association degrees' crisp value, for each of the rules \(R_1\)-\(R_3\), denoted \(h_{crisp}\) is also listed. }
    \centering
    \renewcommand{\arraystretch}{1.2}
    \scalebox{1.1}{
    \begin{tabular}{l | l | l l l l l |l}
    \textbf{R} & \textbf{h} & \(z_{0.2}\) & \(z_{0.4}\) & \(z_{0.6}\) & \(z_{0.8}\) & \(z_{1.0}\) &\( h_{crisp}\) \\ \hline
    % [0.077, 0.086, 0.089, 0.081, 0.073]
    %[0.124, 0.126, 0.128, 0.11, 0.103]
         \multirow{2}{*}{\(R_1\)}&\(L\) & 0.077& 0.086& 0.089& 0.081& 0.073&\multirow{2}{*}{0.097}\\
         &\(U\) & 0.124& 0.126& 0.128& 0.11& 0.103 & \\\hline
         %--------------------------------------------
         %[0.217, 0.239, 0.243, 0.225, 0.213]
         %[0.326, 0.336, 0.331, 0.322, 0.308]
         \multirow{2}{*}{\(R_2\)}&\(L\) & 0.217& 0.239& 0.243& 0.225& 0.213&\multirow{2}{*}{0.274}\\
         &\(U\) &0.326& 0.336& 0.331& 0.322& 0.308 &\\\hline
         %----------------------------------------------
         % [0.057, 0.054, 0.054, 0.056, 0.058]
         %[0.071, 0.065, 0.069, 0.069, 0.072]
         \multirow{2}{*}{\(R_3\)}&\(L\) & 0.057& 0.054& 0.054& 0.056& 0.058 &\multirow{2}{*}{0.063}\\
         &\(U\) & 0.071& 0.065& 0.069& 0.069& 0.072 &\\\hline
    \end{tabular}
    }
    \label{tab:Class_AD}
\end{table}
%---------------------------------------------------------
%---------------------------------------------------------

\noindent The same process can be repeated for each time interval with their respective rules to predict a label for the output. Hence, in this numerical example, there will be three output labels for a total of three time intervals.

%---------------------------------------------
%---------- Classification Step by Step -- END
%---------------------------------------------
%---------------------------------------------------------------------
%---------------------------------------------------------------------
\subsection{Estimating Temporal Trajectories from TXAI Models}
\label{sec:RTM}
%---------------------------------------------------------------------
%-
The temporal trajectories of a dynamic system can be outlined by the TXAI system by making use of the conditional distribution integrated into the TXAI system. The trajectories of a TXAI model is motivated by the work of Filev et al.  \cite{Filev_2013_MarkovModels} that embodies fuzzy transition events defined by joint possibility encompassing the current and future prototypical rules. More specifically, the TXAI system can delineate a rule transition matrix (RTM) which will entail the joint possibility of the rules in present (\(\Delta t\)) and future (\(\Delta t^+\)) time intervals. In mathematical terms, for a total of U rules in time interval \(\Delta t\), and a total of V  rules in time interval \(\Delta t^+\), the RTM can be written as follows \cite{Filev_2013_MarkovModels}:
\begin{equation}
RTM (\Delta t,\Delta t^+) = \begin{bmatrix}
\pi_{11} & ... & \pi_{1N}\\
\rotatebox[origin=c]{90}{...} & \rotatebox[origin=c]{-45}{...}  & \rotatebox[origin=c]{90}{...} \\
\pi_{M1} & ... & \pi_{UV}
\end{bmatrix}
\end{equation}
			     	     
where \(\pi_{cd}\) is the rule transition possibility (RTP) for the \(c^{th}\) rule, \(r_c\), in time interval \(\Delta t\) and the \(d^{th}\) rule, \(r_d\), in time interval \(\Delta t^+\) as given by (\ref{eq:TPelem}).

\begin{align} \label{eq:TPelem}
    \pi_{cd} = \eta_{cd}\times \frac{S_{cd}}{S_{ \Delta t^+}}
\end{align}

where \(\eta_{cd}\) is the joint possibility for the two rules to be prototypical in their respective time intervals, and the ratio \(\frac{S_{cd}}{S_{\Delta t^+}}\) entails the number of times \(r_c\) and \(r_d\) are observed in their respective time intervals with respect to all V rules in \(\Delta t^+\). The following equations, (\ref{eq:eta_TP}) - (\ref{eq:Ratio_TP}), outline how \(\eta_{cd}\) and the ratio \(\frac{S_{cd}}{S_{ \Delta t^+}}\) are computed. 
 
\begin{align} \label{eq:eta_TP}
    \eta_{cd}(r_{c,\Delta t}, r_{d,\Delta t^+}) = \gamma_{c}(r_c, \Delta t) \times \gamma_{d}(r_d, \Delta t^+)
\end{align}
Where \(\gamma\) is computed by applying the t-norm operator (product or minimum type) to the conditional distribution values of the antecedents of a given rule (\(r\)) in a given time interval (\(\Delta t\) or \(\Delta t^+\)); mathematically expressed as shown in equation (\ref{eq:FoO_TP}) for rule (\(r_c\)) in time interval (\(\Delta t\)). The computation of the conditional distribution, \(f\), is previously outlined in Section \ref{sec:TemporalFS} (in particular see (\ref{eq:FoCC})).

\begin{align}\label{eq:FoO_TP}
    \gamma_c(r_{c,\Delta t}) = f_{c}(Ant_{1, r_c}, \Delta t) \times f_{c}(Ant_{2, r_c}, \Delta t)\times ...\times  f_{c}(Ant_{a, r_c}, \Delta t)  
\end{align}

where \(a\) is the total number of antecedents (Ant) of rule \(r_c\). The elements for computing the ratio \(\frac{S_{cd}}{S_{ \Delta t^+}}\) are outlined in (\ref{eq:Ratio_TP}):
\begin{align}\label{eq:Ratio_TP}
    S_{cd} &= \sum r_{c, \Delta t} r_{d, \Delta t^+} \\ \nonumber
    S_{\Delta t^+} &= \sum_{d=1}^V r_{d, \Delta t^+}
\end{align}

where the numerator, \(S_{cd}\), represents the sigma count of the number of times \(r_c\) and \(r_d\) are observed in their respective time intervals, and the denominator, \(S_{\Delta t^+}\), denotes the sigma count of observing all V rules in \(\Delta t^+\).

%---------------------------------------------------
%---------------------------------------------------
\section{Case Study: Time-dependant Occupancy Dataset}
%---------------------------------------------------
%---------------------------------------------------
%---------------------------------------------------
\begin{table*}[htbp]
\caption{The classification problem is exemplified using the proposed Time-dependent eXplainable Artificial Intelligence (TXAI) system with occupancy dataset \cite{Candanedo_2016_OccupancyDataset}. The output for the classification problem predicts the label of whether the room is occupied or not. The time points (\(t_n\)) for calculating the frequency of occurrence are 24 on account of the number of hours in a given day with a total of three corresponding time intervals (\(\Delta t_q\)) of Morning (time \(<\) 11 am), Daytime (11 am \(<\) time \(<\) 7pm), and Evening (time \(>\) 7pm).}
    \centering
    \renewcommand{\arraystretch}{1.35}
    \begin{tabular}{l| l l l l l}
    \textbf{Problem} & \textbf{Input/Output} & \textbf{Feature/Label} & \textbf{Conceptual Labels (CoLs)} & \textbf{N}  & \textbf{Time Intervals, \(\Delta t_q\)}\\ \cline{1-6}
      \multirow{4}{*}{Classification} &
      \multirow{3}{*}{Input} & Temperature & Low, Medium, High & 24 & Morning, Daytime, Evening\\
        & &\cellcolor{gray!10}Light & \cellcolor{gray!10}Low, Medium, High & \cellcolor{gray!10}24 & \cellcolor{gray!10}Morning, Daytime, Evening\\
        & & CO\(_2\) & Low, Medium, High & 24 & Morning, Daytime, Evening\\
        \cline{2-6}
      &Output &\begin{tabular}{l}
\hspace{-.2cm}Occupied   \\
\hspace{-.2cm}Not Occupied 
      \end{tabular} & - &- & Morning, Daytime, Evening\\ \hline
    \end{tabular}
    \label{tab:Class_Problem}
\end{table*}
%-------------------------------------------
%-------------------------------------------
%-------------------------------------------

In this section, a temporal occupancy dataset \cite{Candanedo_2016_OccupancyDataset} is used to exemplify the proposed TXAI system modelling. The occupancy dataset entails measurements of a room along with the time of when the measurement is recorded. In particular, it includes measurements of the room temperature, light, CO\(_2\), and a binary label of whether or not the room is occupied. There are 8,143 data instances in the dataset taken over a period of a few weeks. 

In this work, the dataset \cite{Candanedo_2016_OccupancyDataset} is used for classification problem where TXAI system predicts whether or not the room is occupied based on the room measurements. The inputs of temperature, light, and \(CO_2\) are used to predict whether or not the room is occupied. Three CoLs of Low, Medium, and High are associated with inputs of temperature, light, and CO\(_2\). The primary MF of the CoLs for all inputs are empirically found. The time is discretised at each hour of the day hence a total of \(N= 24\) time points with a total of three time intervals defined at Morning, Daytime, and Evening, as also summarised in Table \ref{tab:Class_Problem}. The z-slices are obtained on locations [0.2, 0.4, 0.6, 0.8, 1.0].  \textcolor{blue}{All aforementioned parameters values are selected so as to reflect the inherent dynamics of the system (such as discretising time at each hour) and to obtain a good enough TXAI model without adding too much computational complexity, for example, the more the z-slices the more accurate the TXAI model would be but at a greater computational cost($p= Q * z_I$ but independent of the data size in each \(\Delta_{t_q}\)).}

%From an implementation point of view, the proposed TXAI systems' complexity is linear with respect to the constant parameters related to the number of discrete time intervals and number of z-slices, i.e. $p=t * z$, but independent of the data size in each time interval, therefore $n=1$, i.e. complexity is linear on $O(p)$. TT2FSs computational complexity is affordable and feasible for nowadays standard off-the-shelf multi-core computer processors, like the ones we used in this study (ARM Neoverse V1). Regarding space complexity TT2FSs are also feasible, as all observed data $n$ can be represented in a multi-dimensional array defined by the number of time intervals and z-slices, therefore complexity is again linear with respect to these parameters, and constant (independent) with respect to the total number of data points.

The conditional distribution for each CoL of every input is computed on the entire dataset. Once the conditional distributions are computed, the learning procedure focuses on the data belonging to each interval. A 10-repeated nested cross-validation procedure is adopted. The dataset is split into a disjoint stratified train, validation and the test set to ensure a random selection of the datasets (train, validation, and test) is not creating any bias in the results. Each repetition, 20\% of the dataset is held out as a test set, and the remaining is used to build the train and validation sets. Train and validation sets are determined in an inner 10-fold procedure, where a fold is used for validation and the rest for training to determine the rule weights. Balanced accuracy and other performance metrics are computed over each validation and test set. 

A rule-base is formed for each time interval. The rules are learned using GA \cite{herrera2008} such that they (rules) attain optimally balanced accuracy on the validation datasets. The GA parameters specification includes the number of generations, set at 20, with each generation having a population of 50. Moreover, the GA is leveraged to find the rules that are prototypical for each time interval. The number of antecedents in each rule can be at most 3 but not more to underpin explainability and hamper model complexity therefore precluding over-fitting. For the same reason, the maximum number of rules in each candidate rule-base for each time interval was limited to 30, although further pruned when its weight (eq. (\ref{eq:RW})) does not surpass a tolerance threshold of 0.001.  

In order to compare the performance of the proposed TXAI system, numerous state-of-the-art classifiers which can both analyse time-series data and/or are explainable have been used. More specifically, for comparison with temporal analysis Long Short-Term Memory (LSTM) \cite{Sherstinsky_2020_LSTM} and Hidden Markov Models (HMM) \cite{Fine_1998_HMM} are used, for comparison with explainable models the standard GT2 based XAI system is used, and for partial explainablility Decision Trees (DT) \cite{Ben-Haim_2010_DT} is used. In addition, a comparison is also made with a temporal convolutional network (TCN) \cite{TCN_github} for comparison with deep learning methods \cite{DeepLearning}. Parametrization and configuration was set to default mode of their respective libraries (Sklearn and Keras). For methods with no modelling with respect to a time component, time is given as an extra input feature. Moreover, the train, validation, and test dataset splits are similar across all methods and for GT2 based XAI in particular, the location of z-slices, and the GA parameters for rule learning are also identical to those of TXAI system. 

%For the explainable systems, the only difference in the rules is that for standard XAI system, time interval (Morning, Daytime, and Evening) will appear as an antecedent in the rules whereas for the TXAI system, the rules are found for each time interval separately. 

\subsection{Results}

%---------------------------------------------------------
%----------------------- comparison with XAI --------------------------
%---------------------------------------------------------
\begin{figure*}
\caption{\small A comparison of the classification prowess of the proposed time-dependent eXplainable artificial intelligence (TXAI) system with numerous state-of-the-art classification systems: temporal convolutional networks (TCN), Long Short-Term Memory (LSTM), Decision Trees (DT), Hidden Markov Models (HMM), and the standard general type-2 (GT2) based XAI system for the classification problem using an occupancy dataset \cite{Candanedo_2016_OccupancyDataset}. a) and b) show the classification metrics of the aforementioned systems on 10 times the 10-fold stratified validation and test dataset respectively. The classification metrics are balanced accuracy (BAcc), recall, precision, and f-score. c) A comparison of the convergence of TXAI with GT2 based XAI system using balanced accuracy for a total of 20 generations with a population of 50 each resulting in a total of 1000 function evaluations.}
\hspace{-.5cm}
    \begin{tabular}{l l l}
    \includegraphics[trim =.3cm .1cm 1.45cm .1cm, clip = true ,scale=.45]{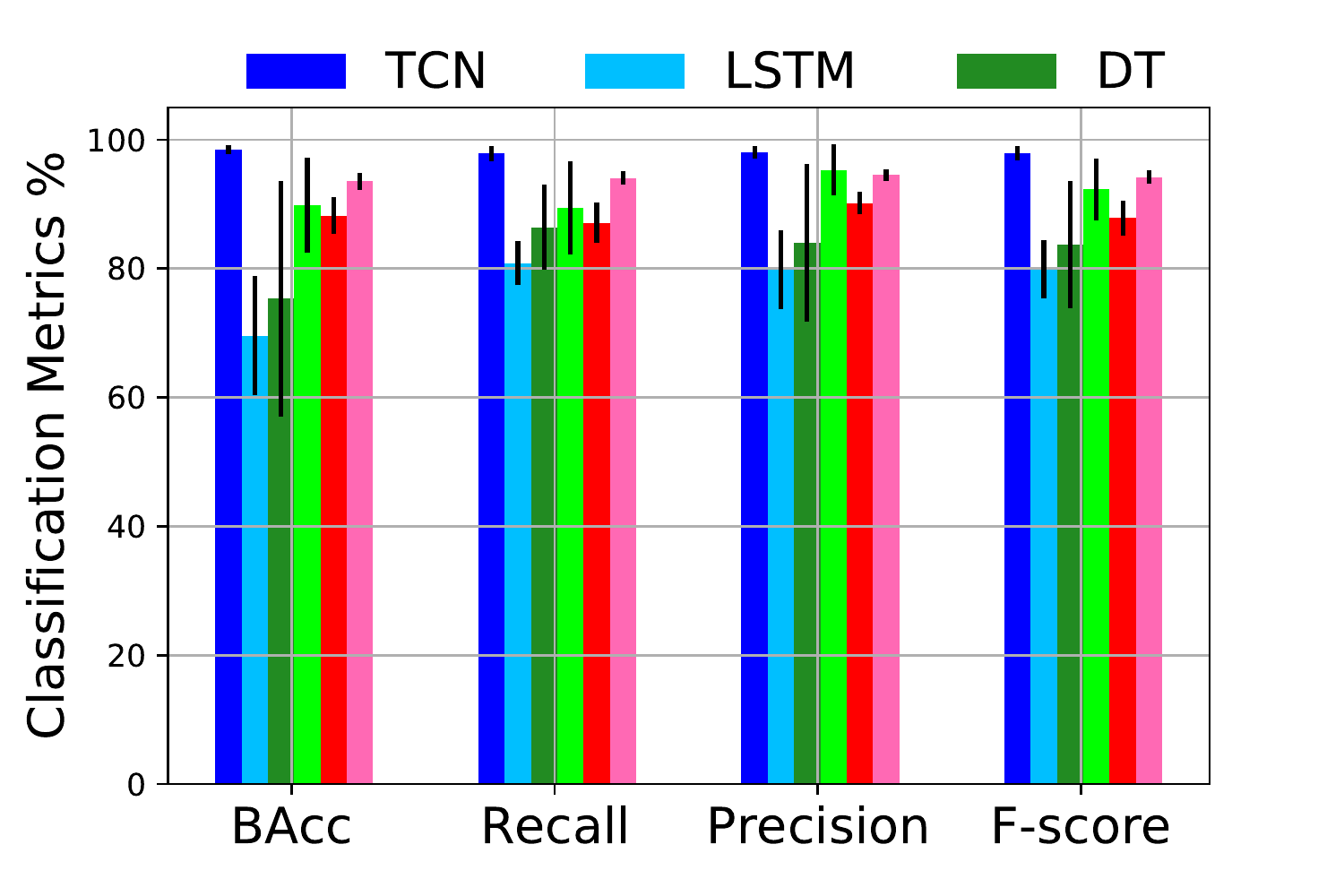}  &
    \includegraphics[trim =.3cm .1cm 1.45cm .1cm, clip = true ,scale=.45]{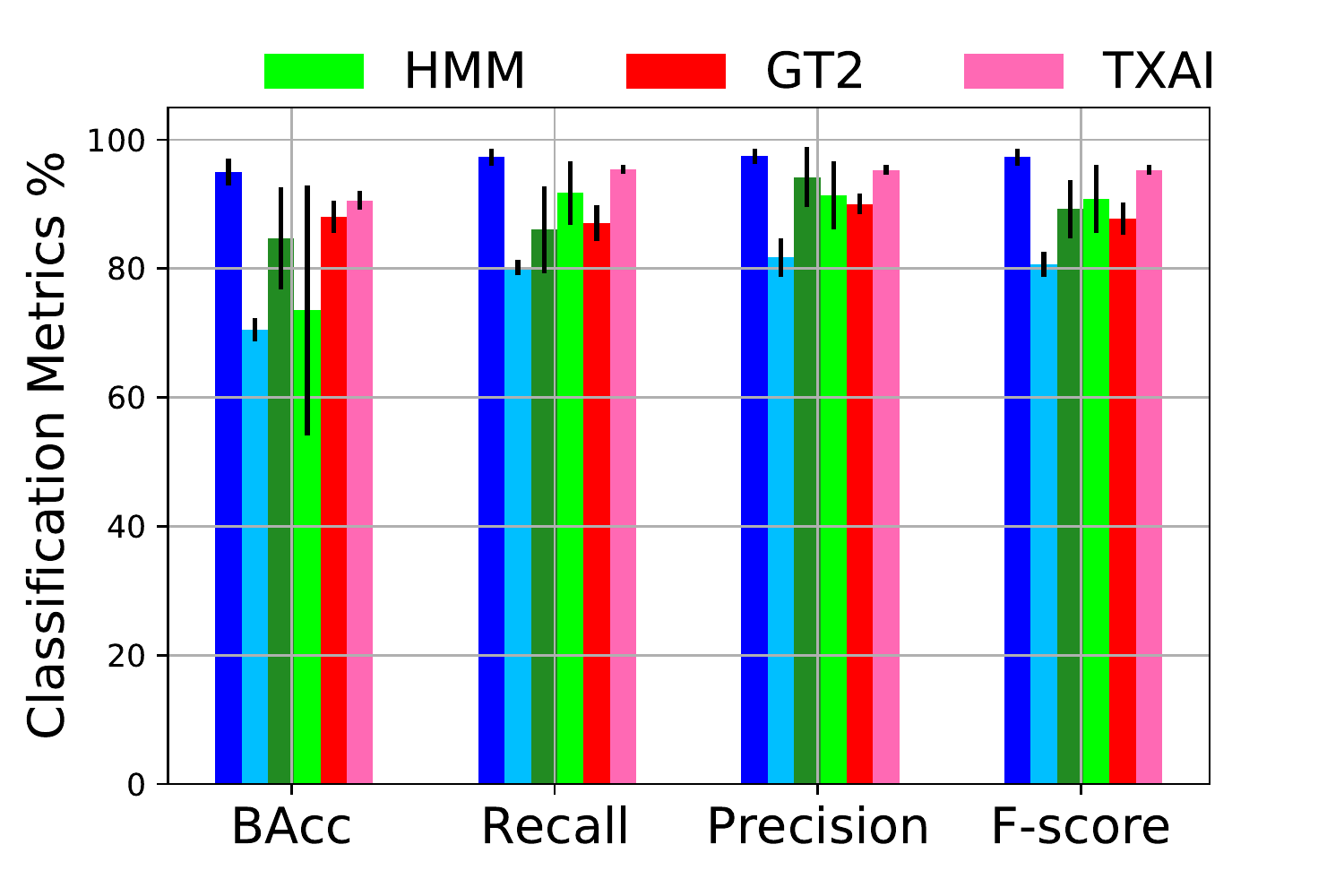}  &
    \includegraphics[trim =.45cm .1cm 1.45cm .1cm, clip = true ,scale=.45]{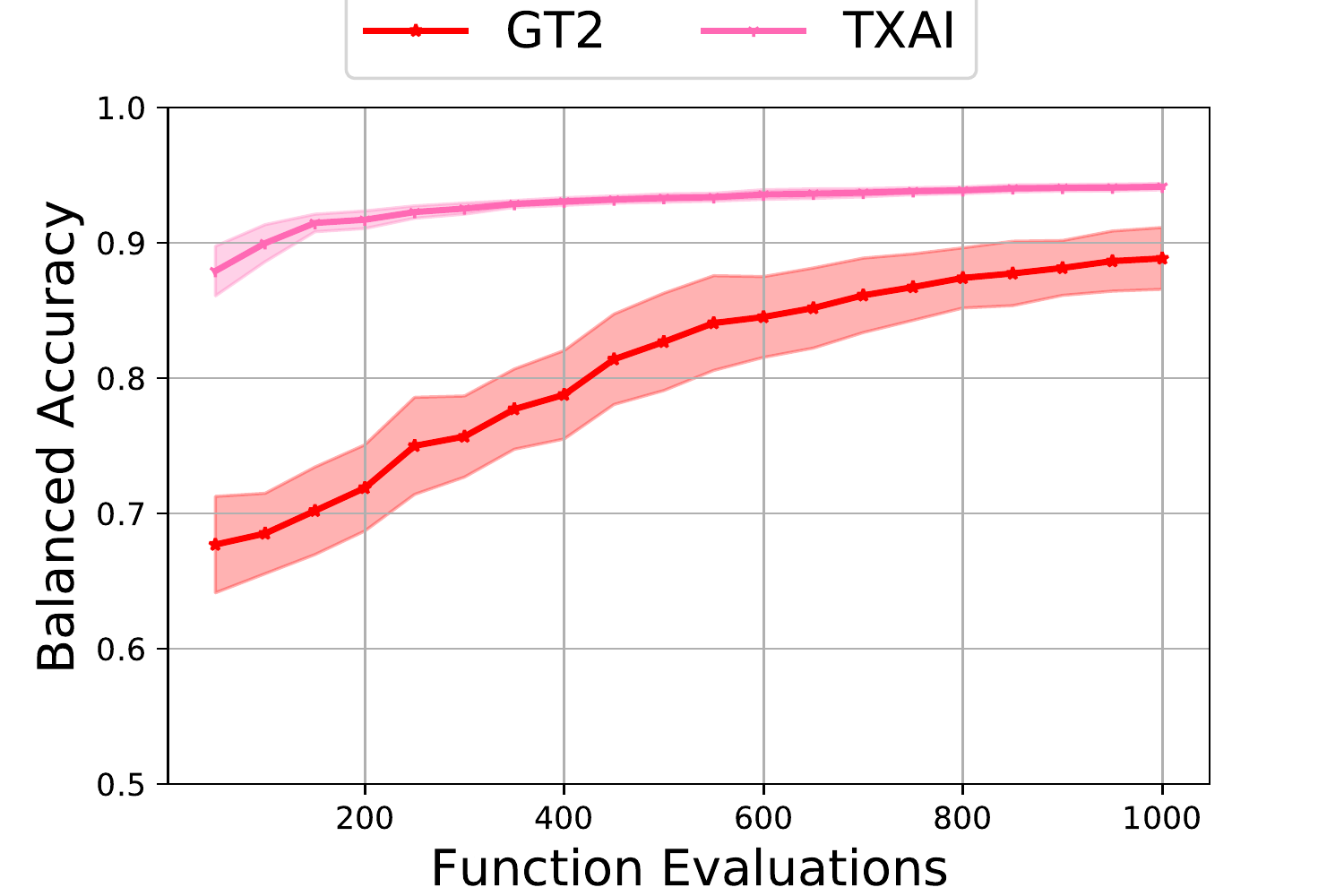} \\
    \multicolumn{1}{c}{\small{a) Validation datasets}} & \multicolumn{1}{c}
    {\small{b) Test datasets}} & \multicolumn{1}{c}{\small{c) Convergence graph}} 
    \end{tabular}
    \label{fig:CompResults}
\end{figure*}
%---------------------------------------------------------
%---------------------------------------------------------
%---------------------------------------------------------
For the classification problem undertaken, using the occupancy dataset, the proposed TXAI system and numerous state-of-the-art classification methods predict whether or not the room is occupied. The mean (and standard deviation) f-score obtained using TXAI system on the 10 test datasets is 95.30\% which is the highest score on the test dataset across all classifiers except TCN. The other classification metrics investigated in this work are balanced accuracy, recall, and precision. A bar plot for the aforementioned classification metrics for both the proposed TXAI and the state-of-the-art AI methods (TCN, LSTM, DT, HMM, GT2 based XAI) on 10 times repeated 10-fold validation and test datasets is shown in Fig. \ref{fig:CompResults} (a) and (b) respectively. In addition, a convergence graph that outlines how the GA optimisation converges with respect to balanced accuracy for both TXAI and GT2 based XAI systems is also shown in Fig.  \ref{fig:CompResults} (c).

The rules outlined by TXAI and  GT2 based XAI systems which are prototypical for whether or not the room is occupied are listed in Table \ref{tab:TXAI_Rules}. For the TXAI system, please note that the rules are found separately for each time interval (Morning, Daytime, and Evening) whereas, for GT2 based XAI system, the time intervals are one of the antecedents of the rules. In general, for both TXAI and GT2 based XAI systems, the rules outline that when the room measurements have higher values, the room is more likely to be occupied, and when the room measurements are on the lower end, the room is more likely to be not occupied. 

For the TXAI system, the temporal trajectories of a time-variant system can also be investigated using the rule transition matrices (RTMs), previously outlined in section \ref{sec:RTM}. The individual RTMs transitioning from one time interval (\(\Delta t\)) to another i.e., from Morning to Daytime, from Daytime to Evening, and from Evening to Morning, represent the joint possibilities of observing a given rule in \(\Delta t^+\) with respect to the rules in  \(\Delta t\). The rules corresponding to the highest RTPs (rule transitioning possibilities) are also joined with lines in the column \textit{RT} (rule transitions) in Table \ref{tab:TXAI_Rules} and illustrated in a schematic in Fig. \ref{Fig:RT_Schematic}.

\begin{table*}[htbp]
    \centering
    \caption{The prototypical rules were obtained by the proposed time-dependent explainable artificial intelligence (TXAI) system for the binary classification problem (room occupied or not) using the occupancy dataset. In the column RT (Rule Transition), the rules with the highest rule transition possibility (RTP) for transitioning from one time interval to another are marked with connecting lines: red line connects the rules with the highest RTP for going from Morning to Daytime, blue line connects the rules with the highest RTP for going from Daytime to Evening, and green lines connects the rules with the highest RTP for going from Evening to Morning of the next day. The numerical values of the corresponding RTPs are also listed. The rules obtained using the standard general type-2 (GT2) explainable artificial intelligence (XAI) system with time as another input are also outlined at the end of the table for comparison purposes. }
    %\rowcolors{2}{gray!25}{white}
    \renewcommand{\arraystretch}{1.58}
    \scalebox{0.8}{
    \begin{tabular}{c c c l  l l l}
    \rowcolor{gray!20}
    \textbf{Method} & \textbf{Time} & \textbf{Rule No.} &\textbf{Rule}  & \textbf{Rule Weight} &  \textbf{Rule Transition (RT)} \\ \hline
    %-----------------------------------------------------------
    %-------------------------------------- TXAI Morning Rules---------------------
    %-----------------------------------------------------------
    \multirow{25}{*}{\rotatebox[origin=c]{90}{Time-dependent eXplainable Artificial Intelligence (TXAI)}} &   \multirow{12}{*}{\rotatebox[origin=c]{90}{Morning}} & 1 &IF Light is High THEN room is Occupied & 0.346 & 
    %-----------------------------------------------------------
    %------- draw the RTPs ---------------------------------------------
    %----------------------------------------------------------
    \multirow{25}{*}{
    \begin{tikzpicture}
    \vspace{-1cm}
    \draw(0,0) -- (0,0);
    % From Morning to Daytime
    \draw [red, ultra thick](0,-4.25) -- (.25,-4.25);
    \draw [red, ultra thick](0.25,-4.25) -- node {\midarrow} (.25,-5.75);
    \draw [red, ultra thick](0.25,-5.75) -- (0,-5.75);
    \node[text = red, rotate=90] at (-0.1,-5) {\large{0.218}};
     % From Daytime to Evening
    \draw [blue, ultra thick](0.35,-5.75) -- (.75,-5.75);
    \draw [blue, ultra thick](0.75,-5.75) -- node {\midarrow} (.75,-10.25);
    \draw [blue, ultra thick](0.75,-10.25) -- (0.35,-10.25);
     \node[text = blue, rotate=90] at (0.4,-8.25) {\large{0.335}};
    % From Evening to Morning
    \draw [green, ultra thick](0.85,-10.25) -- (1.25,-10.25);
    \draw [green, ultra thick](1.25,-10.25) -- node {\midarrowother} (1.25,-4.25);
    \draw [green, ultra thick](1.25,-4.25) -- (0.85,-4.25);
    \node[text = green, rotate=90] at (1.6,-7.25) {\large{0.226}};
    \end{tikzpicture}}\\
    %-----------------------------------------------------------
    %-----------------------------------------------------------
    %-----------------------------------------------------------
    & & \cellcolor{gray!10}2 & \cellcolor{gray!10}IF Temperature is High THEN room is Occupied & \cellcolor{gray!10}0.079\\
    & & 3 &IF CO\(_2\) is Medium THEN room is Occupied & 0.050\\
    & & \cellcolor{gray!10}4& \cellcolor{gray!10}IF CO\(_2\) is High THEN room is Occupied & \cellcolor{gray!10}0.046\\
    & & 5&IF Temperature is High AND Light is High THEN room is Occupied & 0.018\\
    & & \cellcolor{gray!10}6&\cellcolor{gray!10}IF Light is High AND CO\(_2\) is Medium THEN room is Occupied & \cellcolor{gray!10}0.014\\
    & & 7&IF Light is High AND CO\(_2\) is High THEN room is Occupied & 0.012\\
    & & \cellcolor{gray!10}8 &\cellcolor{gray!10}IF Temperature is Medium THEN room is Occupied & \cellcolor{gray!10}0.012\\
    & & 9 &IF Temperature is High AND CO\(_2\) is High THEN room is Occupied & 0.011\\
    & & \cellcolor{gray!10}10 &\cellcolor{gray!10}IF Temperature is Medium AND CO\(_2\) is Medium THEN room is Occupied & \cellcolor{gray!10}0.007\\ \cline{4-5}
    %-----------------------------------------------------------
    & & 11 &IF Light is Low THEN room is Not Occupied & 1.000\\
    & & \cellcolor{gray!10}12 &\cellcolor{gray!10}IF Temperature is Low THEN room is Not Occupied & \cellcolor{gray!10}0.073\\ \cline{2-5}
    %-----------------------------------------------------------
    %--------------------------------------TXAI Daytime Rules---------------------
    %-----------------------------------------------------------
    & \multirow{9}{*}{\rotatebox[origin=c]{90}{Daytime}}  & 1 &IF Light is High THEN room is Occupied & 0.473 & \\ 
    & & \cellcolor{gray!10}2 &\cellcolor{gray!10}IF Temperature is High THEN room is Occupied & \cellcolor{gray!10}0.277 \\ 
    & & 3 &IF CO\(_2\) is High THEN room is Occupied & 0.110 \\
    & & \cellcolor{gray!10}4 &\cellcolor{gray!10}IF Temperature is Medium AND Light is High THEN room is Occupied & \cellcolor{gray!10}0.017 \\
    & &5 &IF Temperature is High AND Light is High AND CO\(_2\) is High THEN room is Occupied & 0.015 \\ \cline{4-5}
    %-----------------------------------------------------------
    & &\cellcolor{gray!10}6 &\cellcolor{gray!10}IF Light is Low THEN room is Not Occupied & \cellcolor{gray!10}1.000& \\
    & &7 &IF CO\(_2\) is Low THEN room is Not Occupied & 0.50& \\
    & &\cellcolor{gray!10}8 &\cellcolor{gray!10}IF Light is Medium THEN room is Not Occupied & \cellcolor{gray!10}0.147\\
    & &9 &IF Temperature is High AND Light is Low THEN room is Not Occupied & 0.011\\ \cline{2-5}
     %-----------------------------------------------------------
    %--------------------------------------TXAI Evening Rules---------------------
    %-----------------------------------------------------------
    & \multirow{4}{*}{\rotatebox[origin=c]{90}{Evening}} & \cellcolor{gray!10}1 & \cellcolor{gray!10}IF Light is High THEN room is Occupied & \cellcolor{gray!10}0.005 \\\cline{4-5}
    %-----------------------------------------------------------
    & & 2 &IF Light is Low THEN room is Not Occupied & 1.000\\
    & & \cellcolor{gray!10}3 &\cellcolor{gray!10}IF Light is Low AND CO\(_2\) is Low THEN room is Not Occupied & \cellcolor{gray!10}0.108 \\
    & & 4 &IF Temperature is High AND Light is Low THEN room is Not Occupied & 0.041 \\ \hline
    %-----------------------------------------------------------
    %--------------------------------------XAI Rules---------------------
    %-----------------------------------------------------------
    \multicolumn{2}{c}{\multirow{5}{*}{\rotatebox[origin=c]{90}{\begin{tabular}{c}
         eXplainable Artificial  \\
         Intelligence (XAI)
    \end{tabular}}}}     & \cellcolor{gray!10}1& \cellcolor{gray!10}IF Light is High AND Time is Daytime THEN room is Occupied & \cellcolor{gray!10}0.580  \\
    &  &2 &IF Light is High AND Time is Morning THEN room is Occupied  & 0.425  \\
    &  &\cellcolor{gray!10}3 & \cellcolor{gray!10}IF Temperature is High AND Time is Daytime THEN room is Occupied  & \cellcolor{gray!10}0.419  \\ \cline{4-5}
    %-----------------------------------------------------------
    &  &4 & IF Light is Low AND Time is Morning THEN room is Not Occupied  & 1.000 \\ 
    &  &\cellcolor{gray!10}5 & \cellcolor{gray!10}IF CO\(_2\) is Medium AND Time is Evening THEN room is Not Occupied  & \cellcolor{gray!10}0.789 \\ \\ \hline
    \end{tabular}}
    \label{tab:TXAI_Rules}
\end{table*}

\subsection{Discussion }

In this work, the proposed TXAI system is used to model an occupancy dataset \cite{Candanedo_2016_OccupancyDataset} for the classification problem of whether or not the room is occupied. For comparison purposes, several state-of-the-art explainable (GT2 based XAI system), partially explainable (DT), and non-explainable methods that can analyse temporal information (LSTM and HMM) as well as TCN are also applied to the aforementioned classification problem. As can be noted from the Fig. \ref{fig:CompResults} (a) and (b), TXAI offers greater classification performance than all classifiers (for e.g. for mean fscore TXAI performs better than LSTM by 18.19\%, DT by 6.81\%, HMM by 4.90\% , GT2 based XAI system by 8.58\% on test datasets) except TCN (for mean fscore TCN performs better than TXAI by 4.67\% on test datasets). However, the TCN classification mechanism is not explainable hence unable to shed light on the prediction of the room occupancy based on input features of Temperature, Light, CO\(_2\), and Time. 

With respect to the comparison with the GT2 based XAI system, the only explainable system apart from the proposed TXAI system, a convergence graph plotted in Fig. \ref{fig:CompResults} (c) also highlights that TXAI system converges (\(\sim\)500 function evaluations) twice as faster than standard GT2 based XAI system (\(\sim\)1000 function evaluations) whilst also yielding higher classification metrics (Fig. \ref{fig:CompResults} (a) and (b)). Moreover, the rules outlined by the explainable systems, TXAI and XAI systems, are listed in Table \ref{tab:TXAI_Rules}, and both systems are in agreement that when the room measurements (Temperature, Light, and CO\(_2\)) have higher values, then the room is likely to be occupied, and when the room measurements are lower, then the room is likely to be not occupied. However, the rules for TXAI also offer greater insight into how the room measurements are interlinked with respect to predicting room occupancy. For example, for the time interval Morning, rule no 5 (see Table \ref{tab:TXAI_Rules}) outlines that if both inputs of Temperature and Light have high values then the room is likely to be occupied. In this regard, rules across time intervals shed light on the intertwined CoLs of the inputs prototypical for decoding the room occupancy.% In contrast, the rules obtained using XAI system are limited in this respect; the XAI system rules do not delineate any relation between the inputs for predicting the room occupancy status. As can be noted from Table \ref{tab:TXAI_Rules}, all rules obtained by XAI system only have one input feature from the room measurements.

%---------------------------------------------------------------------
\tikzstyle{dotted_block} = [draw=black!30!white, line width=1pt, dash pattern=on 1pt off 4pt on 6pt off 4pt, inner ysep=1mm,inner xsep=1mm, rectangle, rounded corners ]
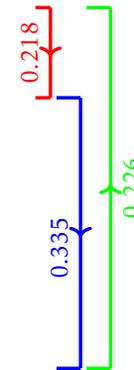
\begin{figure*}[tp]
\caption{A schematic presenting the evolution of the occupancy system based on the rules with the highest rule transition possibilities (RTPs) found by the proposed time-dependent explainable artificial intelligence (TXAI). The rules in two consecutive time intervals with the highest RTPs are linked together to show how the occupancy system is evolving from one time interval to another. A complete list of all the rules delineated by TXAI for the occupancy dataset is enumerated in Table \ref{tab:TXAI_Rules}.}
\hspace{0cm}
\centering
\scalebox{.65}{
\begin{tikzpicture} [node distance =12cm]
 %--------------------------------------------------------
    % Morning
    %--------------------------------------------------------
    \node (Mor1) [block, fill= red!30, text width = 6cm] 
    {\begin{tabular}{l}
            Rule No. 11: IF Light is Low \\
            THEN room is Not Occupied.
    \end{tabular}};
    % %--------------------------------------------------------
    % \node (Mor2) [block, fill= red!30, text width = 9cm, right of = Mor1, inner sep = 0] 
    % {\begin{tabular}{l}
    %      Rule No. 12: IF Temperature is Low   \\
    %      THEN room is Not Occupied.
    % \end{tabular}};
    %--------------------------------------------------------
     \node (Mor) [dotted_block, fit=(Mor1), inner  sep = 3mm, label= {[xshift=0cm]\textbf{Morning}}] 
    {};
    %--------------------------------------------------------
    % Daytime 
    %--------------------------------------------------------
    %--------------------------------------------------------
    \node (Day2) [block, fill= blue!30, text width = 6cm, right of = Mor1, node distance =9cm, xshift = 0cm] 
    {\begin{tabular}{l}
        Rule No. 2 If Temperature if High \\
        THEN room is Occupied.
    \end{tabular}};
    % \node (Day1) [block, fill= blue!30, text width = 9cm , right of = Day2] 
    % {\begin{tabular}{l}
    %      Rule No. 5: IF Temperature is High   \\
    %      AND Light is High AND CO\(_2\) is High \\
    %      THEN room is Occupied.
    % \end{tabular}};
    %--------------------------------------------------------
     \node (Day) [dotted_block, fit=(Day2), inner  sep = 3mm, label= {[xshift=0cm, yshift=0cm]\textbf{Daytime}}] 
    {};
    %--------------------------------------------------------
    % Evening 
    %--------------------------------------------------------
    \node (Eve1) [block, fill= green!30, text width = 6cm, right of = Day2, node distance =3cm, xshift = 6cm] 
    {\begin{tabular}{l}
         Rule No. 2: If Light is Low \\
         THEN room is Not Occupied.
    \end{tabular}};
    %--------------------------------------------------------
     \node (Eve) [dotted_block, fit=(Eve1), inner  sep = 3mm, label= {[xshift=0cm, yshift=0cm]\textbf{Evening}}] 
    {};
     %--------------------------------------------------------
    % % Paths
    \draw[myarrows] (Mor1) -- (Day2) node[midway, yshift = 0.25cm]{\textbf{0.218}};
    \draw[myarrows] (Day2) -- (Eve1)node[midway, yshift = 0.25cm]{\textbf{0.335}};
    \draw[myarrows] (Eve1.south) -- ++(0,-1)-| (Mor1.south)node[midway,xshift = 9cm, yshift = -0.25cm]{\textbf{0.226}};
     %--------------------------------------------------------
\end{tikzpicture}}
\label{Fig:RT_Schematic}
\end{figure*}
%------------------------------------------------------------------

%A possible account for the standard GT2 based XAI system to not offer rules with more input features as antecedents could be that it (rules with more input features as antecedents) yields lower balanced accuracy whereas the GA is constrained to find the highest balanced accuracy hence the omission of the more detailed rules. In this regard, since TXAI system is able to shed light on input feature interaction whilst also performing better at balanced accuracy, with the same GA constraints, implies that the use of conditional distribution (a distinguishing feature from the standard XAI system) to credit the membership values of the inputs does bolster the model performance by the proposed TXAI system. Although it is an empirical finding it warrants further analysis to establish if crediting membership function values with the frequency of occurrence improves the classification prowess of the TXAI system. 

Furthermore, the TXAI systems are also able to shed light on the temporal trajectories of the system being modelled using RTMs, previously outlined in Section \ref{sec:RTM}, and illustrated in Fig. \ref{Fig:RT_Schematic}. The RTPs (rule transition possibilities), which are the elements of the RTMs, represent the joint likelihood of observing a rule in one time interval (rows) and then observing another rule in the next time interval (columns). For example, in the RTM transitioning from Morning to Daytime, the rules with the highest RTP are rule number 12 (for time interval Morning) and rule number 5 (for the next time interval Daytime). For the particular case of the occupancy datasets, the RTMs and the corresponding RTPs outline the trajectory across time as the TXAI model transitions from one time interval to another. In this case, an analysis of the occupancy dataset can be leveraged for the efficient energy management of smart homes using the predictive power of the RTMs \cite{Rocha_2021_SmartHomes}. 

\textcolor{blue}{Indeed, the motivation for developing the TXAI systems is to be able to analyse time-dependent real processes across time. In this regard, conditional distribution integrated within the TXAI model can be used to obtain the RTMs. The RTMs entail the likelihood of observing the transition of a real-life process from one time point to another. %In particular, the authors developed the TXAI system, as an extension of xMVPA \cite{AndreuPerez_2021_xMVPA}, to be able to investigate functional brain development as previously mentioned in Introduction. 
The proposed TXAI system can shed light not only on which rules are prototypical for each of the time intervals but also on the likelihood of observing the rules across the different time points. }%In this sense, the TXAI system would be able to delineate the brain development trajectories crucial for understanding typical and atypical brain development.
%\input{Dynamic_FS_ComprehensionExample}
%---------------------------------------------------------------------
%---------------------------------------------------------------------
\section{Conclusion}
The ability of an explainable system to model a  real-life process in terms of its characteristic features is of paramount significance to inform about the nature of the process. In this regard, XAI systems have proved pivotal for increasing our understanding of numerous complex real-life processes. However, non time-dependent XAI systems are not able to analyse real-life processes across time. This is a critical limitation of standard XAI systems for modelling time-variant real-life processes, especially where time is a defining parameter for the model, i.e., the real-life process behaves differently across time (for example, functional brain development \cite{Kiani_2021_FunctionalBrainReview_UnderReview, AndreuPerez_2021_xMVPA,Kiani_2022_effective,achanccaray_2017}). %For functional brain development analysis it is pertinent to also incorporate associated time information with the acquired brain data to delineate brain developmental trajectories, i.e., how is the brain developing across time. In this regard, our earlier work delineated cortical networks formed for the processing of visual and auditory stimuli in six-month-old infants using an XAI system based on IT2 fuzzy sets called xMVPA (explainable multivariate pattern analysis) \cite{AndreuPerez_2021_xMVPA}. However, xMVPA could only analyse neuroimaging data recorded from infants of the same age (cross-sectional data) and is not capable to analyse longitudinal data (data acquired from infants at different ages such as 3, 6, 9, and 12 months). Infants' longitudinal data analysis can delineate brain development trajectories since it holds information of how the infants' brain is working across different times/ages. It is important to investigate the brain development trajectories as they can inform on typical and atypical brain development, which in turn can be leveraged to inform clinical, educational and social policies. 
To this end, in this work, we propose a new time-dependent XAI system, called TXAI, characterised with time-conditioned distribution for analysing a time-variant real-life process across time. 

The proposed TXAI system can delineate the trajectories of a dynamic, real-life process across time. In addition, a comparison with state-of-the-art AI systems, with varying levels of explainability, manifested that the proposed TXAI performed better than most of the compared AI systems (for e.g. for mean fscore TXAI performs better than LSTM by 18.19\%, DT by 6.81\%, HMM by 4.90\%, GT2 based XAI system by 8.58\% on test datasets) except TCN which is a much more complex, and a black-box method. XAI systems based on standard FLS (e.g., T1, IT2 or GT2) are unable to integrate information relative to the time dimension. More specifically, TXAI system credit the membership value of a fuzzy concept given the fuzzy concept is likely to occur at the time of observation of fuzzy concept using conditional distribution. The conditional distribution is then utilised to investigate the evolution of the process across different time intervals. In this way, the TXAI is able to predict the likelihood of observing prototypical rules of the process across different time intervals. %This is of particular interest for scientists investigating functional brain development, where the brain development changes significantly with respect to time. 
\textcolor{blue}{For future works, the proposed TXAI system can have profound implications to contribute to our understanding of temporal real-life processes, for instance human-centred systems and life sciences. Further, for these future life science studies, we would also endeavour that TXAI entails all ethical concerns accounted for a more fair, and complete TXAI analysis.}\\
\small{\textbf{Acknowledgement}: We would like to gratefully acknowledge \emph{Oracle for Research}, and specially Richard Pitts, and Mike Reilly for their technical assistance and support in the computational resources in Oracle Cloud Resource for this research.}

\bibliography{bibliography}
%---------------------------------------------------------------------
%---------------------------------------------------------------------
\end{document}